\documentclass[sigconf]{acmart}

\AtBeginDocument{%
  }

\usepackage{graphicx}
\usepackage{booktabs} 
\usepackage{hyperref}

\usepackage{amsmath}
\usepackage{amsthm}
\usepackage{algorithm}
\usepackage{algorithmic}
\usepackage{mathtools}
\usepackage[capitalize,noabbrev]{cleveref}
\usepackage{graphicx}
\usepackage{booktabs}
\usepackage{wrapfig}
\usepackage{balance}  
\usepackage{cancel}
\usepackage{lipsum}\usepackage[utf8]{inputenc}
\usepackage{forest}
\usepackage{booktabs}
\usepackage{xspace}
\usepackage{enumitem}
\usepackage{hyperref}
\usepackage{caption}
\usepackage{pifont}
\usepackage{subcaption}
\usepackage{subcaption}
\usepackage{orcidlink}
\usepackage{xcolor,colortbl}
\usepackage{natbib}
\usepackage{enumitem}

\usepackage{orcidlink}
\newcommand{\x}{\mathbf{x}}
\newcommand{\y}{\mathbf{y}}
\newcommand{\oo}{\mathbf{o}}
\newcommand{\z}{\mathbf{z}}
\newcommand{\R}{\mathbb{R}}
\newcommand{\LL}{\mathbb{L}}
\newcommand{\PP}{\mathbb{P}}
\newcommand{\HH}{\mathbb{H}}
\newcommand{\citeauthornum}[1]{\citeauthor{#1}~\cite{#1}}
\usepackage{xcolor}

\definecolor{bg}{rgb}{0.95,0.95,0.95} 

\usepackage{comment}
\usepackage[english]{babel}
\usepackage{hyperref}     

\renewenvironment{proof}{\begin{newproof}}{\end{newproof}\qed}

\theoremstyle{plain}

\theoremstyle{definition}

\theoremstyle{remark}

\copyrightyear{2025}
\acmYear{2025}
\setcopyright{cc}
\setcctype{by-nc-sa}
\acmConference[KDD '25]{Proceedings of the 31st ACM SIGKDD Conference on
Knowledge Discovery and Data Mining V.2}{August 3--7, 2025}{Toronto, ON,
Canada}
\acmBooktitle{Proceedings of the 31st ACM SIGKDD Conference on Knowledge
Discovery and Data Mining V.2 (KDD '25), August 3--7, 2025, Toronto, ON,
Canada}
\acmDOI{10.1145/3711896.3736564}
\acmISBN{979-8-4007-1454-2/2025/08}

\begin{document}

\title{Hyperbolic Deep Learning for Foundation Models: A Survey}

\author{Neil He}
\email{neil.he@yale.edu}
\orcid{0009-0008-3193-2448}
\affiliation{%
  \institution{Yale University}
  \city{New Haven}
  \state{Connecticut}
  \country{USA}
}

\author{Hiren Madhu}
\email{hiren.madhu@yale.edu}
\orcid{0000-0002-6701-6782}
\affiliation{%
  \institution{Yale University}
  \city{New Haven}
  \state{Connecticut}
  \country{USA}
}

\author{Ngoc Bui}
\email{ngoc.bui@yale.edu}
\orcid{0000-0002-4345-6003}
\affiliation{%
  \institution{Yale University}
  \city{New Haven}
  \state{Connecticut}
  \country{USA}
}

\author{Menglin Yang}
\email{menglin.yang@outlook.com}
\orcid{0000-0003-2510-5282}
\affiliation{%
  \institution{Hong Kong University of Science and Technology (Guangzhou)}
  \city{Guangzhou}
  \state{}
  \country{China}
}\authornote{Corresponding author} 

\author{Rex Ying}
\email{rex.ying@yale.edu}
\orcid{0000-0002-5856-5229}
\affiliation{%
  \institution{Yale University}
  \city{New Haven}
  \state{Connecticut}
  \country{USA}
}

\renewcommand{\shortauthors}{Neil He, Hiren Madhu, Ngoc Bui, Menglin Yang, and Rex Ying}

\begin{abstract}
  Foundation models pre-trained on massive datasets, including large language models (LLMs), vision-language models (VLMs), and large multimodal models, have demonstrated remarkable success in diverse downstream tasks. However, recent studies have shown fundamental limitations of these models: (1) limited representational capacity, (2) lower adaptability, and (3) diminishing scalability. 
  These shortcomings raise a critical question: \textbf{ is Euclidean geometry truly the optimal inductive bias for all foundation models}, or could incorporating alternative geometric spaces enable models to better align with the intrinsic structure of real-world data and improve reasoning processes?
  Hyperbolic spaces, a class of non-Euclidean manifolds characterized by exponential volume growth with respect to distance, offer a mathematically grounded solution. These spaces enable low-distortion embeddings of hierarchical structures (e.g., trees, taxonomies) and power-law distributions with substantially fewer dimensions compared to Euclidean counterparts. Recent advances have leveraged these properties to enhance foundation models, including improving LLMs’ complex reasoning ability, VLMs’ zero-shot generalization, and cross-modal semantic alignment, while maintaining parameter efficiency.
  This paper provides a comprehensive review of hyperbolic neural networks and their recent development for foundation models. We further outline key challenges and research directions to advance the field.

\end{abstract}



\maketitle
\section{Introduction}Foundation models, such as large language models (LLMs)~\citep{zhao2023survey}, vision-language models (vLMs)~\citep{zhang2024vision}, diffusion generative models~\citep{cao2024survey}, biological foundation models~\cite{cui2024scgpt, chen2022interpretable}, and large multimodal architectures~\citep{caffagni2024revolution}, have revolutionized machine learning pipelines, enabling unprecedented performance and versatility across a wide variety of tasks and domains~\citep{myers2024foundation, awais2025foundation}. Their widespread success is mainly due to extensive pre-training and robust transfer learning capabilities~\citep{bommasani2021opportunities}. However, despite their transformative impact, current foundation models rely predominantly on Euclidean geometry, which inherently constrains their \textit{representational capacity}, \textit{adaptability}, and \textit{scalability}--properties critical to the success of existing foundation models and for their continued advancement.

As existing foundation models assume linear relationships and uniform distributions among embedded tokens or data points~\citep{gower1985properties}, they have limited representation capabilities. Several works have found that token distributions and real-world data exhibit often hierarchical and scale-free properties~\cite{he2025position, yang2024hypformer,coenen2019visualizing, mettes2024hyperbolic, peng2021hyperbolic}, deviating from the Euclidean assumption. For example, in language, relationships such as entailment and syntax naturally occur in tree-like structures, reflecting conceptual hierarchies~\cite{nickel2017poincare, tifrea2018poincar, le2019inferring}. Social networks and recommendation systems also exhibit hierarchical relationships, where entities form nested communities and user preferences~\cite{yang2021discrete, yang2022htgn, chen2021modeling,yang2022hrcf,yang2022hicf,ma2024harec}. In scientific domains, hierarchical structures emerge in protein-modeling~\cite{villegas-morcillo2021protein,alanis2018latent} and single-cell RNA-seq analysis~\cite{ding2021deep,bhasker2024contrastive}. However, when such hierarchies are embedded into Euclidean space, significant distortions~\citep{linial1995geometry, sala2018representation} arise, especially in low dimension settings, limiting the effectiveness of Euclidean models in capturing these relationships~\cite{krioukov2010hyperbolic, sarkar2011low}.

Furthermore, an essential hallmark of foundation models is their adaptability: a single pretrained model is fine-tuned or transferred across various domains. Many desirable downstream tasks naturally exhibit hierarchical or power-law relationships that are more suitably modeled in geometries with negative curvature~\cite{bai2021modeling, yang2022htgn, kennedy2013hyperbolicity, poleksic2023hyperbolic, tang2023hyperbolic,papadopoulos2010greedy}. However, the restrictive nature of Euclidean space's fixed zero curvature imposes rigidity~\cite{sanborn2024beyond}, thus failing to achieve optimal performance when adapting to tasks that inherently exhibit hierarchical or power-law characteristics~\cite{chen2024hyperbolic}.

Finally, the scalability of foundation models poses increasing challenges due to their immense dimensionality and the explosive growth in computational~\cite{hoffman2022training}. Euclidean geometry exacerbates this issue as it suffers from fundamental distortion bottlenecks and dimensionality-distortion tradeoffs, requiring extraordinarily high-dimensions to represent complex relationships effectively~\cite{he2025position}. Hyperbolic geometry, however, provides an efficient alternative: its unique capability to embed hierarchical structures and power-law distributions with substantially lower distortion means that models can achieve similar or superior representational quality using significantly fewer dimensions~\cite{nickel2017poincare,HNN, sarkar2011low}. This dimensional efficiency translates to reduced computational resources, lower memory footprint, and consequently improved scaling capabilities.

These shortcomings raise a critical question: could alternative geometric spaces enable foundation models to better align with the intrinsic structure of real-world data and reasoning processes? With mathematically provable guarantees for low-distortion embeddings of tree-like structures and power-law distributions~\cite{sarkar2011low,sala2018representation}, hyperbolic geometry has emerged as a compelling alternative to overcome the fundamental limitations posed by Euclidean geometry~\cite{nickel2017poincare,nickel2018learning,HNN,hgcn2019,yang2024hypformer,liu2022enhancing,zhang2025understanding} in recent years; rapid development in hyperbolic deep learning have led to foundation models that leverage hyperbolic representations to substantially improve expressiveness, transferability, and scalability~\cite{desai2023hyperbolic,yang2024hypformer}. 

This survey aims to systematically review recent advancements in explicitly integrating hyperbolic geometry into foundation models. We review hyperbolic building blocks, comprehensively categorize current hyperbolic foundation models based on their learning paradigms, and finally highlight existing challenges that will guide further research. Compared to existing hyperbolic deep learning surveys~\cite{mettes2024hyperbolic, peng2021hyperbolic, yang2022hyperbolicsurvey}, this paper focuses on hyperbolic foundation models, offering a comprehensive review of the technical details of hyperbolic language, vision, and multi-modal foundation models. In contrast, \citeauthornum{mettes2024hyperbolic} focuses on hyperbolic vision models and does not include hyperbolic LLMs; \citeauthornum{peng2021hyperbolic} focuses generally on hyperbolic neural networks and does not include recent developments in foundation model settings; and \citeauthornum{yang2022hyperbolicsurvey} focuses specifically on hyperbolic GNNs.

\begin{figure} 
    \centering
        \centering \includegraphics[width=0.45\linewidth]{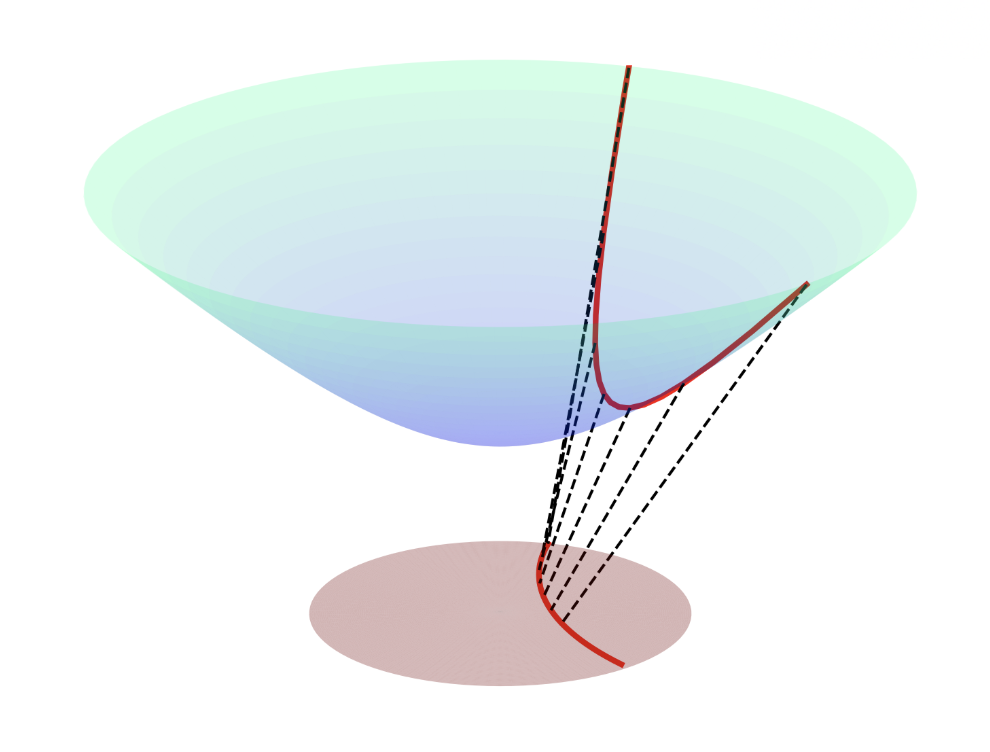}
       \caption{Visualization of $\LL^{K,n}$ (top) and $\PP^{K,n}$ (bottom). The red line shows the \textit{same} geodesic under an isometry.}
       \label{fig:manifold}
       \vspace{-10pt}
\end{figure}

\section{Background}
\subsection{Hyperbolic Geometry}
\textbf{Hyperbolic Space.} Hyperbolic spaces $\HH^{K,n}$ are Riemannian manifolds with constant negative curvature $-K < 0$. There are several models that are isometric, meaning that there is a smooth correspondence between points in different models that preserves distances, angles, and geodesics, allowing for the selection of the most suitable model for a given application. Common models for deep learning include the \textit{Poincaré ball model} $\mathbb{P}^{K, n}$ and the \textit{Lorentz hyperboloid} $\mathbb{L}^{K, n}$. $\PP^{K,n}$ is the sphere of radius $\sqrt{-1/K}$ with a special distance function, and $\LL^{K,n}$ is a $n$-dimensional manifold embedded in $\R^{n+1}$. where each $\x = [x_t,\x_s]^T\in\LL^{K,n}$ is parametrized by its time-like ($\x_t$) and space-like ($\x_s$) dimensions. See \cref{appendix:background} for details.

\textbf{Tangent Space, Geodesics, and M{\"o}bius Operations.} Each point $\x\in$ is associated with a \textit{tengant} space $\mathcal{T}_\x\HH^{K,n}\cong \R^n$, which is a first-order local approximation. \textit{Geodesics} are shortest paths connecting $\x,\y\in\HH^{L,n}$. They induce \textit{exponential maps} $\exp^{K}_\x:\HH^{K,n}\to\mathcal{T}_\x\HH^{K,n}$ and \textit{logarithmic maps} $\log^{K}_\x:\mathcal{T}\HH^{K,n}\to\HH^{K,n}$. \textit{Parallel Transport maps} $PT_{\x\to\y}^K:\mathcal{T}_\x\HH^{K,n}\to\mathcal{T}_\y\HH^{K,n}$ are generalizations of transportation. These functions enable \textit{M{\"o}bius operations}, which are numerical operations computed on the tangent space and lifted back into $\HH^{K,n}$. We denote \textit{M{\"o}bius addition} as $\oplus_K^{PL}$ and \textit{M{\"o}bius multiplication} as $\otimes^{_K}$, which can involve scalers, matrices, or vectors. 

\subsection{Foundation Models}
Foundation models are large neural networks that are pre-trained on vast amounts of data~\cite{bommasani2021opportunities,brown2020language}. These models serve as the backbone of modern AI systems, encompassing LLMs, ViTs, bio-foundation models, and multi-modal architectures. 
The availability of vast amounts of data has led to a widespread adaptation of foundation models in many domains. For example, Large language models such as LLaMa, and DeepSeek~\cite{dubey2024llama3,deepseekv3} have revolutionized natural language processing (NLP). Similarly, vision models like ViT~\cite{dosovitskiy2020image}, multi-modal models like CLIP~\cite{radford2021CLIP}, and biological models like scGPT~\cite{cui2024scgpt} and RNA foundation model~\cite{chen2022interpretable} have shown their effectiveness in their respective fields.

These models are general-purpose architectures. Instead of being designed for a single task, they are trained on diverse, large-scale datasets and fine-tuned for specific applications. This shift from task-specific models to versatile pre-trained architectures has significantly improved the efficiency and accessibility of AI systems. However, traditional foundation models were developed within the Euclidean space, leveraging standard neural network operations such as linear transformations and attention mechanisms, all of which are naturally compatible with the Euclidean framework.

\section{Hyperbolic Basic Neural Operations}\label{sec:building blocks}
In this section, we discuss basic hyperbolic neural operations, which are building blocks for hyperbolic foundation models. We defer the discussion specific to foundation models to \cref{Sec:foundation models}.

\subsection{Basic Operations}\label{sec:hnn}
\textbf{Tangent Space-Based Operations.} Designing hyperbolic neural network operations requires operations to meet manifold constraints while preserving desirable properties. A naive implementation is through the tangent bundle of the manifold, first proposed in HNN~\cite{HNN}, where the intuition is to utilize the fact that the tangent space at any point is Euclidean. These operations extend to all formulations of hyperbolic space. Let $f:\R^n\to \R^m$ be an Euclidean map, such as \textbf{non-linear activation}, \textbf{scalar multiplication}, or \textbf{linear transformation}. Its counterpart as a tangent-space-based-hyperbolic-function~\cite{HNN} is given by $f^{\mathcal{T}, K}:\HH^{K,n}\to\HH^{K.m}$  
\begin{equation}
f^{\mathcal{T}, K}(\x) = \exp_{\oo}^K\left(f\left(\log_\oo^K(\x)\right)\right).
\end{equation} 
It is not difficult to see that concatenating these layers together would cancel out inter-layer $\exp_\oo^K$ and $\log_\oo^K$ maps. 
As a result, \textbf{hyperbolic bias addiction} was proposed, given by $\x\oplus^K_{PL}\mathbf{b}$ (see additions sections). Further refining these operations, HGCN~\cite{hgcn2019} proposed trainable curvature that enables \textit{per-layer varying curvature}, given by $f^{\mathcal{T}}_{K_1, K_2}:\HH^{ K_1,n}\to\HH^{K_2,m}$ for curvature $K_1,K_2$ \begin{equation}
    f^{\mathcal{T}}_{K_1, K_2}(\x) = \sqrt{{K_1}/{K_2}}f^{\mathcal{T}, K_1}(\x)
\end{equation}
and preserves relative distance between points~\cite{hgcn2019}.

\textbf{Fully Hyperbolic Operations.}
Defining neural network operations through the tangent bundle causes several issues, including numerical instability and lack of expressiveness~\cite{chen2021fully, Bdeir2024fully, he2025lresnet}. Many prior works have explored operations that operate directly on the manifold, referred to as \textit{fully hyperbolic}, instead of relying on the local approximations on the tangent space. These operations are typically only available for the Lorentz model, utilizing its unique interactions between the time-like and space-like dimensions.

Linear Lorentz transformations are a combination of \textit{Lorentz rotations} and \textit{Lorentz boosts}~\cite{chen2021fully}, which are not fully covered by the tangent space-based operations. As a result, \textbf{fully hyperbolic Lorentz transformations}~\cite{chen2021fully} were proposed \begin{equation}\label{eq:fnn}
    f^{\mathcal{F}, K}(\x) = \left[\sqrt{\|f(x)\|^2 - 1/K}, f(x)\right]^T,
\end{equation}
where $f:\R^n\to \R^m$ is an Euclidean map. Similar to the case of tangent space-based operations, $f^{\mathcal{F}, K}$ enables a wide range of neural network operations. For instance, a \textbf{fully hyperbolic Lorentzian linear layer}~\cite{chen2021fully} can be defined via a parametrization $f(\x) = f(\mathbf{W}, \x, \mathbf{v}) = \frac{\lambda\sigma\left(\mathbf{v}^T\x+b')\right)}{\|\mathbf{W}h(\x)+\mathbf{b}\|}\left(\mathbf{W}h(\x)+\mathbf{b}\right)$, where $\mathbf{v}\in \R^{n+1}, \mathbf{W}\in\R^{n+1, m}$, $\lambda\in\R$ is a scaling factor, $b\in\R$ and $\mathbf{b}\in\R^{m}$ are bias terms, and $h:\R^{n+1}\to\R^{n+1}$ is an activation function. 

\textbf{Refined Fully Hyperbolic Operations.} $f^{\mathcal{F}, K}$ does not support per-layer change of curvature in the same way as $f^{\mathcal{T}}_{K_1, K_2}$. Hypformer~\cite{yang2024hypformer} proposed \textit{Hyperbolic Transformation with Curvatures}, given by a map $f^{HTC}_{K_1, K_2}:\LL^{ K_1,n}\to\LL^{K_2,m}$ via \begin{equation}
 f^{HTC}_{K_1, K_2}(\x) = \sqrt{{K_1}/{K_2}}\left[\sqrt{\|f(\x)\|^2 - 1/K_1}, f(\x)\right]^T,
\end{equation}
where $f(\x) = f(\x;\mathbf{W},\mathbf{b}) = \mathbf{W}\x + \mathbf{b}$ is a linear layer with bias. Similar to $f^{\mathcal{T}}_{K_1, K_2}$, $f^{HTC}_{K_1, K_2}$ preserves relative distance~\cite{yang2024hypformer}. Hypformer further proposed a series of mappings, \textit{Hyperbolic Readjustment and Refinement with Curvatures}, given by $f^{HRC}_{K_1, K_2}:\LL^{ K_1,n}\to\LL^{K_2,n}$ \begin{equation}
     f^{HRC}_{K_1, K_2}(\x_s) = \sqrt{{K_1}/{K_2}}\left[\sqrt{\|f(\x_s)\|^2 - 1/K_1}, f(\x_s)\right]^T,
\end{equation}
where $f$ can be layer normalization, activation, dropout, etc. 

\textbf{Hyperbolic Multiclass Logistic Regression.} Multiclass logistic regression (MLR) is commonly used to perform multiclass classification in Euclidean space. MLR predicts the probability of vector $\x$ being in class $k$, which is $\mathcal{P}(y=k; \x)\propto e^{v^\R_k(\x)}$, where $v^\R_k(\x) = \langle \mathbf{a}_k, \x\rangle - b_k$ defines the boundary hyperplane. Hyperbolic MLR finds ways to redefine the linear transformation $v$. HNN~\cite{HNN} defined the equivalent operation based on Poincar{\'e} hyperplanes. However, this formulation introduces $2n$ parameters instead of $n+1$ as in the Euclidean case. Through a reparametrization trick, HNN++~\cite{HNN++} reduced the number of parameters from $2n$ to $n+1$.

\textbf{Hyperbolic Fully Connected Layers.}
    The fully connected layer is the building block in many models, such as convolutional layers~\cite{oshea2015introduction}. For the Lorentz model, HCNN~\cite{Bdeir2024fully} proposed using $f^{\mathcal{T},K}$ without normalization, which is equivalent to $f^{HTC}_{K,K}$. An extension is to use the general version $f^{HTC}_{K_1, K_2}$ as the fully connected layer. For the Poincar{\'e} Ball model, HNN++~\cite{HNN++} proposed formulating the fully connected layer using the linear transformation $v_k$ from their MLR construction, instead of relying on tangent space operations. 

\textbf{Hyperbolic Residual Connection and Additions.} Several studies have proposed how to perform residual connections in hyperbolic space, which can be viewed as generalizations of the addition of hyperbolic vectors. The naive implementation in Poincar{\'e} ResNet~\cite{van2023poincar} adopts the parallel transport-based bias addition from earlier to the residual connection, given as $\x\oplus_{PL}^K\y$.
However, this formulation has several disadvantages, such as mapping errors and numerical instabilities~\cite{he2025lresnet}. HCNN proposed to sum the space-like dimension of the two vectors and then compute the time-like dimension of the sum. However, this formulation has an ambiguous geometric meaning and leaves performance to be desired~\cite{he2025lresnet}. LResNet proposed adapting the Euclidean weighted sum to hyperbolic space based on the Lorentzian centroid to resolve these limitations. For $\x,\y\in\LL^{K,n}$, their sum is $\x\oplus^{\mathcal{L}}_K\y = \alpha_{w_x,w_y}\x + \beta_{w_x,w_y}\y$ such that \begin{equation}\label{lrn}
    \begin{split}
        &\alpha_{w_x,w_y} = {w_x}/{\sqrt{-K} |\| w_x\mathbf{x} + w_y f(\mathbf{x}) \|_{\mathcal{L}}|}\\
        &\beta_{w_x,w_y} = {w_y}/{\sqrt{-K} |\| w_x\mathbf{x} + w_y f(\mathbf{x}) \|_{\mathcal{L}}|}
    \end{split},
\end{equation}
where $\alpha_{w_x,w_y}, \beta{w_x,w_y}$ are weights parametrized by $w_x,w_y>0$.

\textbf{Hyperbolic Normalization.} Layer and Batch normalization are essential for neural networks~\cite{ioffe2015batch, ba2016layer}, especially in Transformer-based foundation models~\cite{vaswani2017attention}. Hypformer adapted layer normalization to hyperbolic space, where the operation is defined through $f^{HRC}_{K_1,K_2}$ from earlier. \citeauthornum{lou2020differentiating} proposes using the Fr{\'e}chet mean for batch normalization, which is computationally expensive~\cite{van2023poincar}. Instead, Poincar{\'e} ResNet~\cite{van2023poincar} and HCNN~\cite{Bdeir2024fully} propose computing the batch mean with Poincar{\'e} and Lorentzian midpoints instead, improving computation efficiency. HELM~\cite{he2025helm} further proposed hyperbolic root-square mean (RSM) normalization and proved it to be invariant in both forward and backward passes.

\textbf{Hyperbolic Midpoints.} Hyperbolic attention mechanisms are typically implemented through hyperbolic midpoint operations. ~\citet{lou2020Frechet} proposed using the Fr{\'e}chet mean, which does not have a closed form. HAN~\cite{gulcehre2019hyperbolicAT} proposed using the Einstein midpoint in the Klein model. \citeauthornum{law2019lorentzian} proposed the Lorentzian mean \begin{equation}
    \mathrm{Mid}^\LL_K(\x_1,\ldots,\x_m; \{v_i\}) = \frac{\sum_{j=1}^{m} \nu_j \mathbf{x}_j}{\sqrt{-K} \left\lVert \sum_{i=1}^{m} \nu_i \mathbf{x}_i \right\rVert |_\mathcal{L}|},
\end{equation}
where $\nu_j>0$ are weights, which has a Poincar{\'e} equivalent~\cite{ungar2008gyrovector} \begin{equation}
    \mathrm{Mid}_K^\PP (\x_1,\ldots,\x_m; \{v_i\}) = \frac{1}{2} \otimes_K \frac{\sum_{i=1}^N v_i\lambda^{K}_{x_i} \mathbf{x}_i}{\sum_{i=1}^N |v_i|\left(\lambda^{K}_{x_i} - 1\right)}.
\end{equation}
For the latter three, HNN++~\cite{HNN++} previously showed that all three midpoints are in fact equivalent under the usual isometries.

\textbf{Concatenation and Splitting.} Concatenating and splitting coordinates are vital operations for multi-head attention mechanisms in foundation models~\cite{vaswani2017attention}. HNN~\cite{HNN} proposed hyperbolic concatenation but assumed that it would be followed by a linear layer and requires $N-1$ M{\"o}bius additions~\cite{HNN++}. HNN++~\cite{HNN++} then proposed concatenating hyperbolic vectors with controlled variance, denoted as $\mathrm{Cat}^\PP_K\left(\mathbf{X}\right)$ where$\x_i \in\PP^{K, n_i}$ 
HAEGAN~\cite{qu2022lorentzian} proposed direct concatenation on the Lorentz model, denoted as $\mathrm{Cat}^\LL_K(\mathbf{X})$.
Splitting operations are defined as the inverse of the two above. 

\textbf{Hyperbolic Attention Mechanisms.} 
HAN~\cite{gulcehre2019hyperbolicAT} is the first to propose hyperbolic attention mechanisms. HAN  first computes the queries $\mathbf{Q} = \{\mathbf{q}_i\}$, keys $\mathbf{K} = \{\mathbf{k}_i\}$, and values $\mathbf{V} = \{\mathbf{v}_i\}$ in $\R^n$, lifting the embeddings to $\LL^{-1, n}$, then projecting into the Klein model and compute the hyperbolic attention output based on hyperbolic distance, and finally mapping the vectors back into $\R^n$.

HNN++~\cite{HNN++} proposed Poincar{\'e} self-attention for $\mathbf{Q}, \mathbf{K}, \mathbf{V}\in \PP^{n, K}$ \begin{equation}\label{hnn++}
    \mathrm{att}_i^{\PP, K}(\mathbf{Q}, \mathbf{K}, \mathbf{V}) = \mathrm{Mid}_K^\PP(\mathbf{v_1}, \ldots, \mathbf{v}_N; \{\alpha_{i,j}\}_{j=1}^N),
\end{equation}
where $\alpha_{i,j} = g(f(\mathbf{q_i}, \mathbf{k}_j))$ with $g(x) = e^x$ and $f(\x,\y) = -\eta d_\PP(\x,\y)-\zeta$ similar to HAN, which is equivalent to softmax activation given the property of $\mathrm{Mid}_K^\PP$, The feature for each head is concatenated using $\mathrm{Cat}_K^\PP$. FNN~\cite{chen2021fully} then proposed Lorentz self-attention: \begin{equation}\label{fnn}
    \mathrm{att}_i^{\LL, K}(\mathbf{Q}, \mathbf{K}, \mathrm{V})=\mathrm{Mid}_K^\LL(\mathbf{v_1},\ldots,\mathbf{v}_N; \{\alpha_{i,j}\}),
\end{equation}
where $\alpha_{i,j} = \frac{\exp\left(-d^2_\LL(\mathbf{q}_i, \mathbf{v}_j)\right)}{\sum_{\ell}\exp(d^2_\LL(\mathbf{q}_i, \mathbf{v}_\ell))}$ is the softmax similarity score. The multihead Lorentz attention is $f^{\mathcal{F}, K}(\left [\mathrm{\mu_1,\ldots, \mu_N}\right])$ where $\mu_i = \mathrm{att}_i^{\LL,K}$.

Hypformer~\cite{yang2024hypformer} proposed hyperbolic linear attention. For $\mathbf{Q},\mathbf{K},\mathbf{V}$ in $\LL^{K_1,n}$, let $\mathbf{Q}', \mathbf{K}', \mathbf{V}'$ be $\phi(\mathbf{Q}_s), \phi(\mathbf{K}_s), \phi(\mathbf{V}_s)$ respectively, where $\phi(\x) = \frac{\|\tilde{\x}\|}{\|\tilde{\x}^p\|}\tilde{x}^p$ where $\tilde{e}=\mathrm{ReLU}(e)/t$ with parameters $t,p>0$. The linear attention mechanism, where $\mathbf{Z}=\frac{\mathbf{Q}'(\mathbf{K}'^T\mathbf{V}')}{\mathbf{Q}'(\mathbf{K}'\mathbf{1})} + f^{\mathcal{F}}_{K_1,K_2}(\mathbf{V}_s)$, is \begin{equation}
\mathrm{LiAtt}_{K_1,K_2}^\LL(\mathbf{Q}, \mathbf{K}, \mathbf{V})=\sqrt{{K_2}/{K_2}}\left[\sqrt{\mathbf{Z}-1/K_2}, \mathbf{Z}\right]^T.
\end{equation}

HELM~\cite{he2025helm} proposed hyperbolic Multi-Head Latent Attnetion (HMLA). By projecting the hyperbolic keys and values to low-dimensional latent spaces, HMLA only requires storing the latent key and value vectors. As a result, HMLA significantly reduces the size of KV-cache during generation as compared to hyperbolic self-attention mechanisms. By projecting the query vectors as well, HMLA also reduces the active memory during training.

\textbf{Hyperbolic Positional Encoding.} FNN~\cite{chen2021fully} proposed a primitive hyperbolic positional encoding method by adding it as a bias terms, assuming it is followed immediately by a linear layer. Hypformer~\cite{yang2024hypformer} proposed the positional encoding of input vector $\x$ to be $\mathrm{pos}_K^\LL(\x) = \mathrm{Mid}_K^\LL (\x, f^\mathcal{F}_{K_1,K_2}(\x); \{1,\epsilon\})$
where $\epsilon>0$ is a parameter. HELM~\cite{he2025helm} proposed hyperbolic rotary positional encoding (HoPE) and proved similar theoretical guarantees as the Euclidean RoPE, such as long-range decay and ability to learn complex relations~\cite{su2021roformer, barbero2025rope}. Given $\x_i\in\LL^{n,K}$ the $i$-th token, the operation is \begin{equation}
    \mathrm{HoPE}({\mathbf{x}_i}) = \begin{bmatrix}
        \sqrt{\|\mathbf{R}_{i,\Theta}({\mathbf{x}_i})_s\|^2-1/K},&  \mathbf{R}_{i,\Theta}({\mathbf{x}_i})_s
    \end{bmatrix}^T
\end{equation}
where $\mathbf{R}_{i,\Theta}$ is the usual Euclidean RoPE rational matrix.

\subsection{Hyperbolic Neural Network Models}
In this section, we briefly introduce advancements in hyperbolic neural network models. 
These models are building blocks for the foundation models we would introduce in \cref{Sec:foundation models}.

\textbf{Hyperbolic RNNs and MLPs.}
A primitive implementiation of hyperbolic RNNs and MLPs is with the tangent space-based operation $f^{\mathcal{T}}_{K_1,K_2}$~~\cite{HNN}. 
The \textit{fully hyperbolic} versions of these models can be implemented using $f^{\mathcal{F}, K_1}$, e.g., the fully hyperbolic Lorentzian linear layer from FNN~\cite{chen2021fully}. A refined version can be implemented using $f^{HTC}_{K_1,K_2}$ and $f^{HRC}_{K_2,K_3}$ as in Hypformer~\cite{yang2024hypformer}. 
Currently, there does not exist a fully hyperbolic model in the Poincar{\'e} Ball model, as existing works have all relied on the tangent space operation $f^{\mathcal{T}}_{K_1, K_2}$ for implementing hyperbolic activation function in this space.

\textbf{Hyperbolic CNNs and ResNets.}
Several works have extended CNNs and ResNets to hyperbolic space~\cite{HNN++,van2023poincar, Bdeir2024fully}. Poincar{\'e} ResNet~\cite{van2023poincar} proposed Poincar{\'e} convolutiona layers to develop the first hyperbolic ResNet in $\PP^{K,n}$, with a residual block using a Poincar{\'e} batch normalization layer, $f^{\mathcal{T},K}$ as activation with $f = \mathrm{ReLU}$, and $\oplus^{PL}_K$ as the residual connection. HCNN~\cite{Bdeir2024fully} developed the \textit{fully hyperbolic convolutional layer} in the Lorentz model. 
The residual block is built using Lorentzian batch normalization, ReLU implemented equivalent to $f^{HRC}_{K,K}$, and the space-addition residual connection. LResNet~\cite{he2025lresnet} developed hyperbolic ResNets with a more stable and expressive residual connection method $\oplus^\mathcal{L}_K$. 

\textbf{Hyperbolic GNNs.}
As the negative curvatures of hyperbolic spaces make them particularly suitable for learning on tree-like structures, hyperbolic graph neural networks (GNN) achieve significant improvements over their Euclidean counterparts in learning representations on tree-like graphs~\cite{hgcn2019, lgcn, liu2019HGNN, zhu2020gil, dai2021H2H, zhang2021hyperbolic, chen2021fully}. Earlier works relied on the tangent space. For instance, both HGCN~\cite{hgcn2019} and LGCN~\cite{lgcn} defined neighborhood aggregation by projecting to $\mathcal{T}_\oo\HH^{K,n}$, aggregating through the usual Euclidean means, then lifting back to $\HH^{K,n}$. Later works defined neighborhood aggregation through midpoint methods, such as using $\mathrm{Mid}^\LL_K$~\cite{chen2021fully} or mapping Lorentz embeddings to the Klein model and computes the Einstein mean~\cite{dai2021H2H}. HGAT~\cite{zhang2021hyperbolic} extended graph attention networks to hyperbolic space through tangent space attention and aggregation. GIL~\cite{zhu2020gil} proposed to combine both Euclidean and hyperbolic embeddings to leverage geometric information in both spaces.

\textbf{Hyperbolic MoEs.} GraphMoRE~\cite{guo2025graphmore} proposed a graph mixture-of-curvaure (MoE) model in which each expert resides in a separate Riemannian manifold, including hyperbolic space, then aligning the experts through tangent-space-based operations for task-specific distance computation. HELM~\cite{he2025helm} proposed mixture of curvature expert (MiCE) module for hyperbolic LLMs, in which each expert operate on distinct curvature spaces to better reflect the curvature distributin in token embeddings~\cite{he2025position, he2025helm}. The experts' outputs are then combined through manifold operations before passing to the next Transformer block.

\begin{table*}[]
\caption{Summary of hyperbolic foundation models organized by their associated modality. Additional properties such as model architecture type, geometric mode, and associated hyperbolic space formulation are also shown. 
}\label{fig:taxonomy}
\resizebox{0.75\textwidth}{!}{\begin{tabular}{@{}ccccc@{}}
\toprule
\textbf{Architecture} & \textbf{Method} & \textbf{Modality} &  \textbf{Geometric Mode}   & \textbf{Manifold} \\ \midrule
Transformer, Recursive Transformer  & HAN~\cite{gulcehre2019hyperbolicAT} & Text, Graph  & Hybrid & $\mathbb{K}$\\
Transformer  & HNN++~\cite{HNN++} & Text  & Tangent Space & $\mathbb{P}$\\
Transformer  & FNN~\cite{chen2021fully} & Text   & Fully Hyperbolic & $\LL$    \\ 
Transformer  & H-BERT~\cite{chen2024hyperbolic} & Text   & Fully Hyperbolic & $\LL$    \\
Transformer, Graph Transformer  & Hypformer~\cite{yang2024hypformer} & Text, Graph, Image   & Fully Hyperbolic & $\LL$ \\
Fine-Tuning & HypLoRA~\cite{yang2024hyplora} & Text   & Hybrid & $\LL$ \\

LLM & HELM~\cite{he2025helm} & Text   & Fully Hyperbolic & $\LL$ \\
\midrule\midrule
Vision Transformer  & Hyp-ViT~\cite{ermolov2022hyperbolic} & Image   & Hybrid & $\LL, \PP$ \\
Vision Transformer  & HVT~\cite{fein-ashley2024hvt} & Image   & Tangent Space & $\PP$ \\
Vision Transformer  & LViT~\cite{he2025hypercore} & Image   & Fully Hyperbolic & $\LL$ \\
MoCo & HCL~\cite{Ge2022HyperbolicCL} & Image   & Hybrid & $\PP$ \\
SimCLR/RoCL & RHCL~\cite{yue2023hyperbolic} & Image & Hybrid & $\PP$\\
\midrule\midrule
CLIP & MERU~\cite{desai2023hyperbolic} & Text, Image   & Hybrid & $\LL$ \\
BLIP & H-BLIP-2~\cite{mandica2024hyperbolic} & Text, Image   & Hybrid & $\PP$ \\
CLIP & HyCoCLIP~\cite{pal2025compositional} & Text, Image & Hybrid & $\LL$\\
CLIP & L-CLIP~\cite{he2025hypercore} & Text, Image & Fully Hyperbolic & $\LL$\\

\bottomrule
\end{tabular}}
\end{table*}

\begin{figure} 
    \centering
        \centering
       \includegraphics[width=0.80\linewidth]{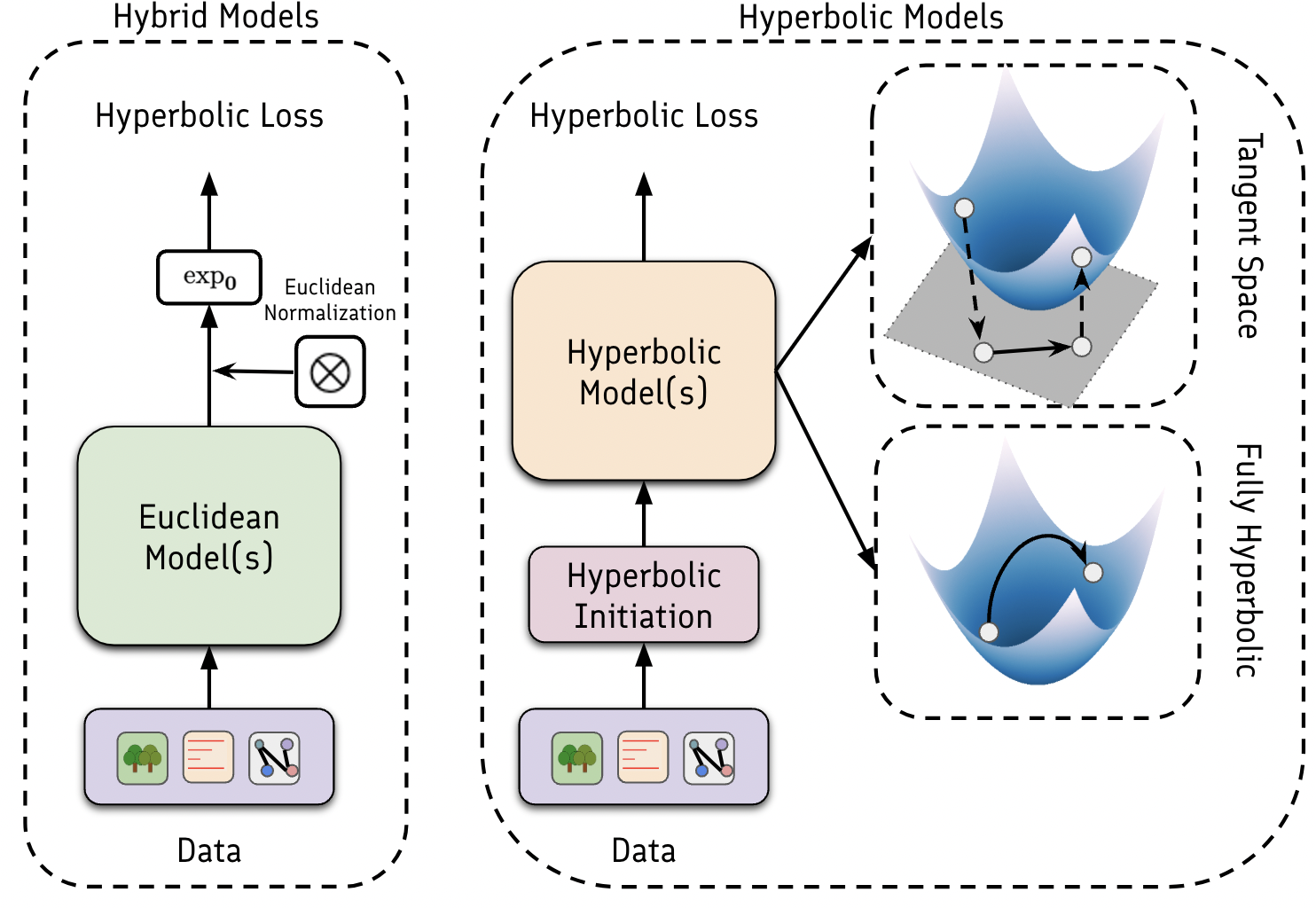}
       \caption{Geometry modes of hyperbolic foundation models. Hybrid models uses hyperbolic and Euclidean modules, and hyperbolic models use only the former. The latter has two types: \textit{fully hyperbolic}, which perform every operation directly on $\HH^{K,n}$; and \textit{tangent space} that performs operations through the tangent space at the origin. Hyperbolic initiation is typically lifting the dataset to the manifold with $\exp_\oo^K$.}
       \label{fig:geo_mode}
       \vspace{-10pt}
\end{figure}

\section{Hyperbolic Foundation Models}\label{Sec:foundation models}
In this section, we discuss recent developments in hyperbolic foundation models, including hyperbolic Transformers and LLMs, hyperbolic vision models, and hyperbolic multi-modal foundation models.
In~\cref{fig:taxonomy}, we organize the models by their modality and provide a summary of model properties, such as geometric mode, the model architecture, and the applicable model of hyperbolic space. See \cref{fig:geo_mode} for an illustration of the geometric modes.

\subsection{Hyperbolic Transformers and LLMs}\label{sec:transformer}
\begin{figure} 
    \centering
        \centering
       \includegraphics[width=0.65\linewidth]{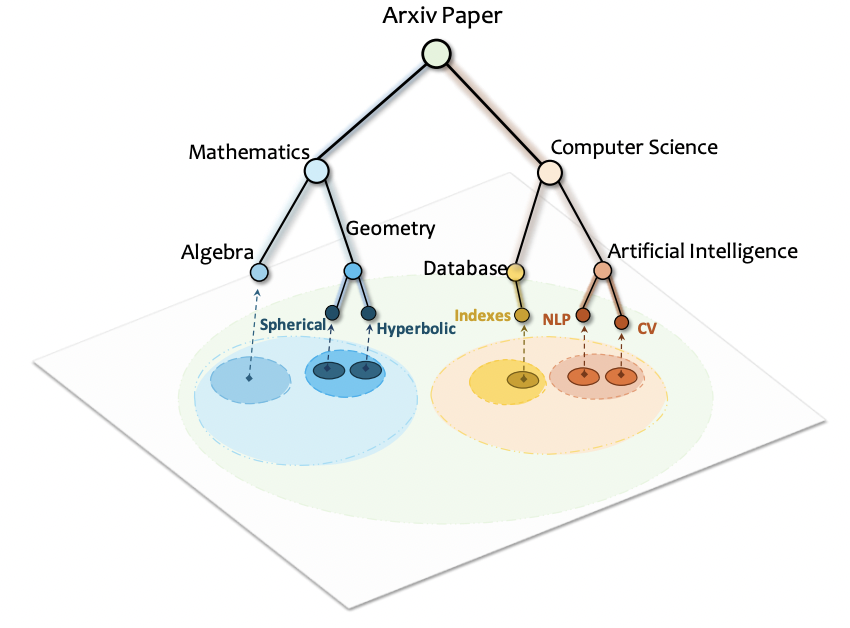}
       \caption{Topics and words can be grouped in a dendrogram to show underlying tree-like concept hierarchies in the data. Image courtesy of~\citet{yang2024hypformer}.}
       \label{fig:hypformer}
       \vspace{-10pt}
\end{figure}

Several works have adapted the Euclidean Transformer~\cite{vaswani2017attention} to hyperbolic spaces~\cite{HNN++, yang2024hypformer, gulcehre2019hyperbolicAT, chen2021fully}, leveraging inherent vocabulary and concept hierarchies in datasets, as shown in \cref{fig:hypformer}. In this section, we discuss the relevant  models and results from these works. See \cref{sec:building blocks} for details regarding the relevant operations.

\textbf{Hyperbolic Transformers.} HAN~\cite{gulcehre2019hyperbolicAT} is the first to propose hyperbolic Transformer models, with methods mentioned in \cref{sec:building blocks}. By replacing the attention layer in Euclidean Transformer models to the Lorentz-Klein formulation of their own, HAN's hybrid Transformer outperformed Euclidean equivalents on both language and graph texts, including tasks on neural machine translation, graph link prediction, and shortest-path length prediction. 

HNN++~\cite{HNN++} then developed the first non-hybrid Transformer using \cref{hnn++}. The architecture also uses hyperbolic components such as $FC_K^\PP$ for feature projection and feedforward layers instead of Euclidean ones. However, the model is missing several components, such as positional encoding and layer normalization. To this end, HNN++ tested the performance of a Poincar{\'e} set Transformer on the task amortized clustering of a mixture of Gaussians (MoG) for both Euclidean and hyperbolic distributions. They found that the hyperbolic set Transformer learns the mixture of hyperbolic Gaussians better than Euclidean counterparts, demonstrating its effectiveness in modeling hierarchical distributions.

FNN~\cite{chen2021fully} then proposed a \textit{fully hyperbolic} Transformer in Lorentz space with \cref{fnn}. The FNN Transformer model uses $f^{\mathcal{F},K}$ for feature projection and feed-forward layers, with positional encoding and residual connection performed added the bias term, where they assumed that each of these operations is followed by a fully connected layer. However, FNN lacked layer normalization layers. On neural machine translation tasks, FNN outperformed Euclidean Transformers, as well as the hybrid HAN model and tangent-space-based HNN++ model. Additionally, FNN found that the HAN model better captures syntax structure than Euclidean Transformers, with their fully hyperbolic Transformer outperforming both. The results shows that as
the model becomes more hyperbolic, the ability to
learn structured information becomes stronger.
Based on FNN, H-BERT~\cite{chen2024hyperbolic} developed and trained larger-scale hyperbolic pretrained BERT models, finding them to outperform Euclidean BERT models on a variety of tasks.

Hypformer~\cite{yang2024hypformer} proposed the first fully hyperbolic Transformer that filled the gaps by developing several essential modules that were previously missing or limited. By optionally combining the Transformer encoder with a hyperbolic GNN, Hypformer's hyperbolic linear attention enabled efficient and effective hyperbolic graph Transformers for large-scale graphs~\cite{yang2024hypformer}, where previous hyperbolic Transfomers fail as they require quadratic time complexity attention layers. Whether incorporating the GNN or not, Hypformer models outperformed Euclidean Transformer models on a variety of graph, image, and text tasks.

\textbf{Hyperbolic Fine-Tuning.} HypLoRA~\cite{yang2024hyplora} recently extended the Euclidean LoRA~\cite{hu2022lora} method for fine-tuning to hyperbolic space. Given learned weights $W\in\R^{m,n}$, Euclidean input vector $\x^H\in\R^n$, learned matrices $A\in \R^{r,n},B\in\R^{m,r}$ the output of HypLoRA is \begin{equation}
    \z^E = W\x^E+\log_\oo^K\left(\mathrm{LLR(B,A, \exp_\oo^K(\x^E))}\right),
\end{equation}
where we have $\mathbf{LLR}(B, A, \mathbf{x}^{H}) =
\left( \sqrt{\|\mathbf{B} \mathbf{y}_s^{H} \|_2^2 + K}, \mathbf{B} \mathbf{y}_s^{H} \right)$ with $\mathbf{y}^{H} =
\left( \sqrt{\|\mathbf{A} \mathbf{x}_s^{H} \|_2^2 + K}, \mathbf{A} \mathbf{x}_s^{H} \right)$. HypLoRA can be applied to fine-tune any Euclidean pre-trained model, showing improvement over models fine-tuned with Euclidean LoRA over math reasoning benchmarks.

\begin{figure} 
    \centering
        \centering
       \includegraphics[width=0.60\linewidth]{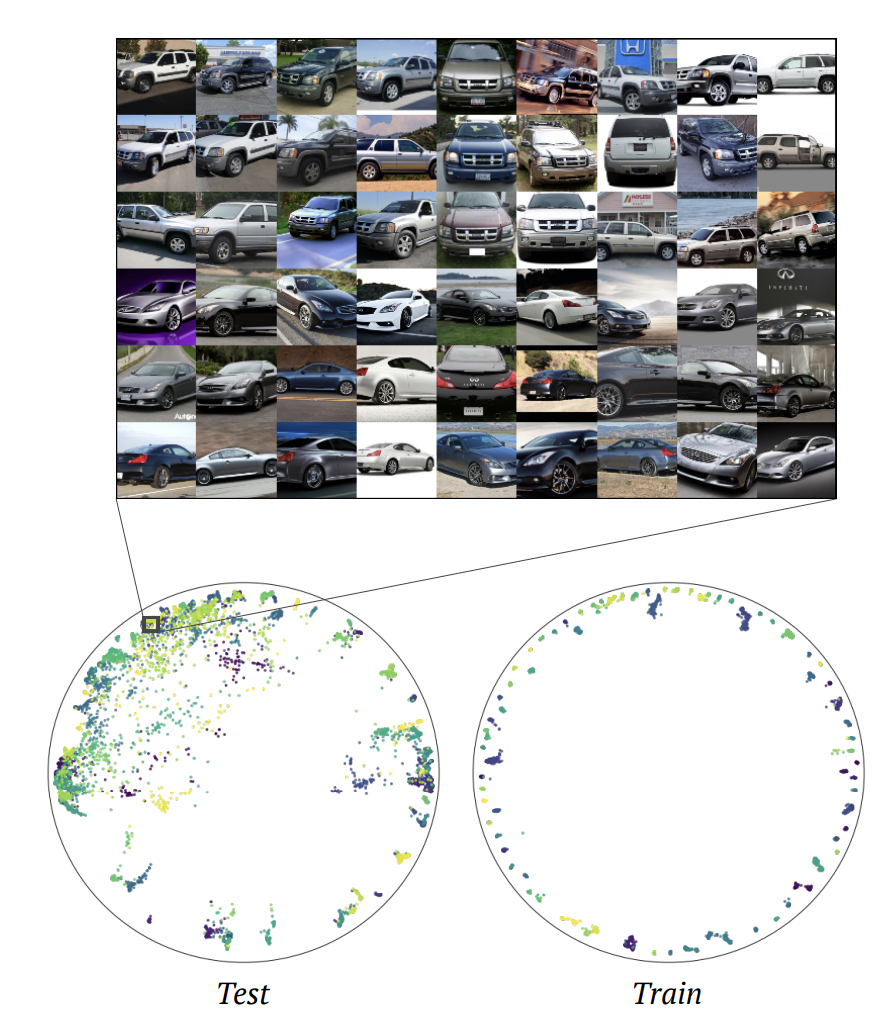}
       \caption{Embeddings of hyperbolic vision transformers push towards the boundary of the Poincar\'e ball based on their labels (colors) and capture latent hierarchical relations. Image courtesy of~\citet{ermolov2022hyperbolic}.}
       \label{fig:hyp-vit}
       \vspace{-10pt}
\end{figure}

\textbf{Hyperbolic LLMs.} HELM~\cite{he2025helm} is the first to develop and train two variants of hyperbolic LLMs: HELM-MiCE, which utilizes the MiCE module better reflect the geometric variation in token distribution~\cite{he2025position}, and HELM-D, a dense model akin to Euclidean models such as LLaMA~\cite{dubey2024llama3}. HELm models integrated HMLA to reduce KV-cache's memory footprint during generation. When trained on the same datasets, HELM models outperformed Euclidean LLMs in billions-of-parameters scale on several benchmarks.
 
\subsection{Hyperbolic Vision Foundation Models.} 
We detail in this section efforts incorporating hyperbolic geometry into vision foundation models through loss functions and modules.

\textbf{Hyperbolic ViTs.} Hyp-ViT~\cite{ermolov2022hyperbolic}, proposed the hyperbolic equivalent of the \textit{cross-entropy loss} to incorporate hyperbolic geometry into vision Transformers (ViT). given as \begin{equation}
    L_{CE}(\z_i,\z_j) = -\log\left(\frac{\exp(-d_\HH(\z_i,\z_j)/\tau)}{\sum_{k=1}^B \exp(-d_\HH(\z_i,\z_k)/\tau)}\right),
\end{equation}
where $B$ is the batch size. Hyp-ViT then trains a hybrid vision Transformer by passing the data through a Euclidean ViT encoder, then projecting the Euclidean embedding into hyperbolic space and computing the hyperbolic cross-entropy loss. The hybrid Hyp-ViT models outperform both Euclidean ViTs and hybrid spherical-Euclidean models~\cite{ermolov2022hyperbolic}. \cref{fig:hyp-vit} shows a projection of the learned embeddings, where
classes are grouped towards the boundary and their latent hierarchical neighborhoods.

HVT~\cite{fein-ashley2024hvt} is the first to build hyperbolic ViTs without relying on Euclidean layers in the Poincar{\'e} Ball model. HVT defined essential operations using the tangent space method $f^{\mathcal{T}, K}$, including linear layer, layer normalization, ReLU activation, layer scaling, residual connection, and attention propagation. The attention score is defined similarly to previous works on hyperbolic Transformers. HVT also proposed learnable positional encoding and a \textit{geodesic regularization} term to stabilize training.

HyperCore~\cite{he2025hypercore} built the first to build full hyperbolic ViTs that do not rely on Euclidean or tangent-space-based operations, which they referred to as LViT. Through Lorentzian convolutional layers, LViT proposed fully hyperbolic patch embedding. HyperCore then adopted HypLoRA to fully hyperbolic settings to build the first fully hyperbolic fine-tuning pipeline for ViTs.

\textbf{Hyperbolic SimCLR and RoCL.} RHCL~\cite{yue2023hyperbolic} incorporates hyperbolic geometry into self-supervised contrastive foundation model learning in two ways. The first adapts SimCLR~\cite{chen2020simple} to hyperbolic space by lifting the Euclidean embedding to hyperbolic space and performing a hyperbolic supervised contrastive loss given as \begin{equation}
    L_{sup}(\x, P)=\frac{-1}{|P|}\sum_{p\in P}\frac{e^{-d_\HH(\x,\x_p)/\tau}}{\sum_{i\in A(x)}e^{-d_\HH(\x,\x_i)}/\tau},
\end{equation} where $P$ is the set of all positive labels and $A(x)$ is the augmented instances excluding $x$. The second adapts Robust Contrastive Learning (RoCL) models~\cite{kim2020adversarial} to learn robust representation. The model first generates label-free adversarial examples using instance-wise adversarial attacks, obtaining $\y=\x+\delta$. The model then trains to maximize the similarity between the clean examples and their instance-wise perturbation, on top of the existing supervised contrastive learning objective, given as \begin{equation}
    L_{robust}(\x,P) = L_{sup}(\x,P\cup\{\y\}) + \lambda L_{sup}(\y, P),
\end{equation}
where $\y$ is a positive label in the first term and as $\x$ in the second.

\textbf{Hyperbolic MoCo.} HCL~\cite{Ge2022HyperbolicCL} adapted MoCo-v2~\cite{chen2020improved}, OCL~\cite{xie2021unsupervised}, and DenseCL~\cite{wang2021dense} models to hyperbolic space for learning object-centric scene hierarchy. HCL extracts a tree-like hierarchy $T=(V,E)$ from image inputs, where the vertices are object boxes, scene regions, and the set of images. There is an edge between $u,v\in V$ if $u\subseteq v$ or $v\subseteq u$. Then the proposed model computes a hybrid contrastive loss where the Euclidean and hyperbolic losses each have their own backbones and momentum encoders. The Euclidean loss is the same as MoCo~\cite{he2020momentum}. For the hyperbolic loss, the model takes in input a scene region $u$, an object region $v$. The final loss is the sum of the two, where the hyperbolic loss is \begin{equation}
    L_{h-moco}(u, v) = -\log\frac{e^{-d_\HH(u,v)/\tau}}{e^{-d_\HH(u,v)/\tau} + \sum_{w\in N_u}e^{-d_\HH(u,w)/\tau}},
\end{equation}
where $\tau$ is the temperature and $N_u=\{w:(u,w)\not\in E\}$.
\subsection{Hyperbolic Multi-modal Foundation Models}
In this section we highlight exciting works that developed hyperbolic contrastive learning methods for foundation models.

\begin{figure} 
    \centering
        \centering
       \includegraphics[width=0.70\linewidth]{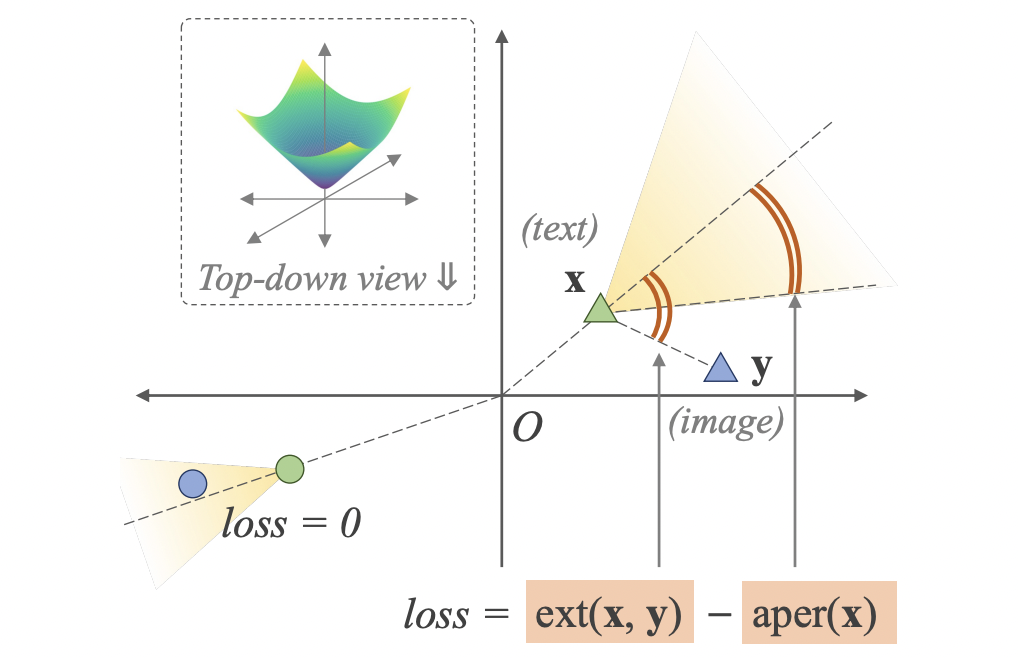}
       \caption{Visualization of hyperbolic entailment cone. This gives rise to hyperbolic entailment loss $L_{ent}$, which pushes image embeddings inside an imaginary cone projected by the text embeddings.  Loss is zero if the image embedding is already inside the cone. Image curtesy of~\citet{desai2023hyperbolic}}.
       \label{fig:meru}
       \vspace{-20pt}
\end{figure}

\textbf{Hyperbolic CLIP.} MERU~\cite{desai2023hyperbolic} was the first to adapt CLIP models~\cite{radford2021CLIP} to hyperbolic space by lifting embeddings from Euclidean models to $\LL^{K,n}$ and computing the hyperbolic contrastive loss~\cite{radford2021CLIP}. Give a batch of size $B$ of image-text pairs and any $j$th instance, its text and image embeddings, $\x_j,\y_j$, form a \textit{positive} pairs, where the rest form \textit{negative} pairs. The \textit{hyperbolic contrastive loss} is  \begin{equation}
    L_{cont}(\x_j,\y_j) = -\frac{1}{2}\log\frac{e^{-d_\HH(\x_j,\y_j)/\tau}}{\sum_{i\ne j}^B e^{-d_\HH(\x_j,\y_i)/\tau}} -\frac{1}{2}\log \frac{e^{-d_\HH(\x_j,\y_j)/\tau}}{\sum_{i\ne j}^B e^{-d_\HH(\x_i,\y_j)/\tau }}
\end{equation}
for temperature $\tau$. MERU also defines the notion of entailment loss in hyperbolic space, similar to prior works~\cite{ganea2018hyperbolic, le2019inferring}. However, MEUR was the first to apply these concepts to incorporate hyperbolic geometry into CLIP models. Hyperbolic entailment loss can be formulated through hyperbolic entailment cones, which is defined through the half-aperture for each origin $\x$: \begin{equation}
    \mathrm{aper}(\x) = \sin^{-1}\left(2\gamma\left({\sqrt{-\frac{1}{K}}\|\x_s\|}\right)^{-1}\right),
\end{equation}
where $\gamma>0$ is a constant that sets boundary behavior near the origin. Then given the entailment cone at $\x$, the exterior angle between $\x$ and the cone is given as \begin{equation}
    \mathrm{ext}(\x,\y)=\cos^{-1}\left(\left(y_t-\frac{x_t}{K}\langle\x,\y\rangle_\mathcal{L}\right)\left({\|\x_s\|\sqrt{\left(\frac{-1}{K}\langle \x,\y\rangle_\mathcal{L}\right)^2-1}}\right)^{-1}\right).
\end{equation}
Intuitively, if $\mathrm{ext}(\x,\y)<\mathrm{aper}(\x)$, then $\y$ is in the interior of the entailment cone and does not need to be penalized. The entailment loss is thus $L_{ent}(\x,\y) = \max(0, \mathrm{ext}(\x,\y) - \mathrm{aper}(\x))$. \cref{fig:meru} shows a visualization of the loss and hyperbolic entailment cone. 
The MERU final loss is then computed as $L_{cont} + \lambda L_{ent}$. 

HyCoCLIP~\cite{pal2025compositional} recently explored hyperbolic CLIP models that leverage more hierarchies than individual image-text pairings. Instead, they consider the hierarchy obtained by decomposing each image-text pairing into multiple object boxes each described by a noun. Given image-text pair $\x_j,\y_j$ and box descriptions $\x^{box}_j,\y^{box}_j$, the \textit{hierarchical compositional contrastive} loss is given as \begin{equation}
   \begin{split}
        &L_{hCC}(\x_j,\y_j,\x_j^{box},\y_j^{box}) \\&= \frac{1}{2}(L_{const}(\x_j,\y_j) + L^*_{const}(\x_j^{box}, \y_j)+L^*_{const}(\y_j^{box}, \x_j))
   \end{split},
\end{equation}
where $L^*_{const}(\x_k,\y_k) = -\frac{1}{2}\log \frac{e^{-d_\HH(\x_k,\y_k)/\tau}}{\sum_{i\ne k}^B e^{-d_\HH(\x_i,\y_k)/\tau}}$. Similarly, the \textit{hierarchical compositional entailment} loss is given as \begin{equation}
    \begin{split}
        &L_{hCE}(\x_j,\y_j, \x_j^{box}, \y_j^{box}) \\&=  L_{ent}(\x_j,\y_j) + L_{ent}(\x_j^{box},\y_j^{box}) + L_{ent}(\x_j^{box},\x_j) + L_{ent}(\y_j^{box},\y_j).
    \end{split},
\end{equation}
The model architecture for HyCoCLIP is similar to that of MERU, with the difference being replacing the loss with $L_{hCC} + \lambda L_{hCE}$. 

HyperCore~\cite{he2025hypercore} recently developed L-CLIP, a fully hyperbolic CLIP model in Lorentz space, where image encoder is LViT and the text encoder is a fully hyperbolic Transformer, demonstrating the potential for building fully hyperbolic multi-modal models to further leverage the data's hierarchial structures.

\textbf{Hyperbolic BLIP.} H-BLIP-2~\cite{mandica2024hyperbolic}, by proposing a hyperbolic-Euclidean hybrid BLIP-2 model, incorporates hyperbolic geometry into multi-modal foundation models on the scale of billions of parameters. Similar to CLIP models, a BLIP-2 model contains an image and a text encoder~\cite{li2023blip2}, with an additional adapter model called the Q-Former. The image encoder maps input images to image embedding $\x$. The image embedding is then passed through the Q-Former to interact with fixed query vectors through cross-attention, obtaining query embeddings $\mathbf{E}=\{\mathbf{e}_j\}_{j=1}^N$. In parallel, the text encoder encodes the caption in text embedding $\y$, where the text is aligned via cosine distance with the most similar query embedding. H-BLIP-2 proposes two ways to adapt this alignment to hyperbolic space. The first is using Poincar{\'e} distance. Let $\mathbf{E},\y$ now represent the hyperbolic embedding after lifting the Euclidean embeddings to $\PP^{K,n}$ through $\exp_\oo^K$. Then the hyperbolic alignment loss is given by $
    L_{align}(\mathbf{E}, \y)=\min_{i=1,\ldots,N}d_\PP(\mathbf{e}_j,\y)$.
However, H-BLIP-2 found that using the above loss pushes the embedding to the boundary of the Poincar{\'e} ball, causing ambiguity when applied to both positive and negative samples. To stabilize hyperbolic BLIP-2 models, a second alignment loss was proposed to apply the Euclidean cosine similarity loss to the hyperbolic embeddings.

\subsection{Tools for Hyperbolic Foundation Models}
Several tools have significantly simplified implementing and training hyperbolic foundation models. Libraries such as Geoopt~\cite{kochurov2020geoopt} and Geomstats~\cite{geomstats} implement basic manifold operations and Riemannian optimizers that simplified training hyperbolic foundation models. HyperCore~\cite{he2025hypercore} is a recently developed library that developed simple-to-use hyperbolic modules that offer more intuitive ways for implementing hyperbolic foundation models, enabling models such as LViT, L-CLIP, and HELM models.

\section{Future Directions and Challenges}
In this section, we discuss several potential future directions as well as their associated challenges.

\textbf{Fully Hyperbolic Pre-trained Models.} Pre-trained models have become a cornerstone of modern training pipelines, providing rich, generalizable representations that enable state-of-the-art transferable learning across diverse downstream tasks~\cite{awais2025foundation, bommasani2021opportunities, devlin2018bert, kapoor2023societal}. While some works have explored fully hyperbolic pre-trained models~\cite{chen2024hyperbolic,he2025helm}, most works still only consider Euclidean models~\cite{yang2024hyplora, ge2023hyperbolic, desai2023hyperbolic, pal2025compositional, yue2023hyperbolic, mandica2024hyperbolic}. While HELM pre-trained hyperbolic LLMs with one-billion parameters, they still do not compare to Euclidean models in terms of the size of the model and training dataset. There lack pre-trained hyperbolic vision models as well. As a result, current works do not fully leveraging the representation power of hyperbolic space. Pre-training hyperbolic models at the scale of Euclidean foundation models could lead to more general hyperbolic representations for downstream tasks. 

\textbf{Parameter-efficient Foundation Models.}
Hyperbolic space can capture hierarchical relationships and scale-free distributions in low-dimensional settings~\cite{sarkar2011low, chen2021fully, HNN++}. As Euclidean foundation models' performance scales inverse exponentially w.r.t parameter count~\cite{hoffman2022training}, hyperbolic foundation models present the exciting potential for more favorable scaling by compressing geometric information~\cite{kiani2024hardness}. This could lead to more parameter-efficient foundation models and provide substantial performance growth at very high dimensions where the growth of Euclidean models diminish. 

\textbf{Efficient and Intuitive Model Training.}
While libraries such as HyperCore~\cite{he2025hypercore} greatly simplifies implementing hyperbolic foundation models, training hyperbolic foundation models comes with its unique challenges. For instance, it is common for prior works to utilize separate optimizers for Euclidean and hyperbolic parameters~\cite{chen2021fully, yang2024hypformer, he2025lresnet}, which is not supported by current foundation models libraries such as DeepSpeed~\cite{rasley2020deepspeed}. This is especially a concern when training hyperbolic foundation models, due to lacking a Riemannian equivalent of the AdamW optimizer widely used to train Euclidean foundation models. Libraries such as FlashAttention~\cite{dao2022flashattention} also do not support hyperbolic attention mechanisms.

\textbf{Hyperbolic Retrieval Augmented Generation.} 
While LLMs have transformed the field of NLP, they struggle with knowledge cutoffs~\cite{cheng2024dated} and hallucinations~\cite{huang2025survey}, leading to inaccurate and out-of-context responses. Retrieval-augmented generation (RAG) has emerged as a powerful approach to improve the factual precision and contextual relevance of LLMs by incorporating external knowledge~\cite{edge2024graphrag, gao2023retrieval, lewis2020retrieval}. Traditional Euclidean RAG systems are not ideal for capturing the hierarchical structures found in real-world text corpora. Hyperbolic retrieval modules, which leverage the hierarchical and scale-free properties of hyperbolic space, could provide a more effective mechanism for document retrieval in knowledge-intensive tasks. The Hyp-GraphRAG model from HyperCore~\cite{he2025hypercore} provides a proof of concept for implementing such models in practice.
Developing fully hyperbolic retrieval pipelines—including hyperbolic nearest neighbor search, ranking mechanisms, and generative architectures—could lead to more structured, accurate, and computationally efficient retrieval-augmented generation systems.

\textbf{Geometric Insights.}
While extensive works exist for hyperbolic foundation models, several essential operations' geometric interpretations remain unclear. For instance, fully hyperbolic operations typically compute the time-like dimension after performing transformations on the space-like dimension. While linear layers and RoPE implemented in this manner have clear geometric meaning~\cite{chen2021fully,he2025helm}, the geometric interpretation for other operations remains under-explored. Whether the fully hyperbolic operation can be found for $\PP^{K,n}$ is unclear as well. Additionally, due to the manifold's un-compactness and negative curvature, current works in hyperbolic diffusion models lack theoretical investigation on their theoretical guarantee as in the case for Euclidean and compact manifolds~\cite{lee2022convergence, debortoli2022riemannian}. These insights could enhance our understanding and potentially lead to more effective and efficient methods. 

\section{Conclusion}
In this article, we present a comprehensive review of current methods for incorporating hyperbolic geometry in foundation model settings. We outline the existing challenges faced by Euclidean foundation models and summarize recently developed hyperbolic basic neural operations, which are building blocks for hyperbolic foundation models. Building on top of this, we discuss the technical details in incorporating hyperbolic geometry into LLMs and Transformers, vision foundation models, and multi-modal foundation models. At the end, we highlight open challenges and propose promising avenues for future investigation. All together, with the surge in foundation models and their limitations with hierarchical datasets, this survey offers a valuable overview for researchers, driving new ideas for future endeavors.

\begin{acks}
This work was supported in part by the National Science Foundation (NSF) IIS Div Of Information \& Intelligent Systems 2403317. We also gratefully acknowledge support in part from the Silicon Valley Community Foundation, an Amazon research award, the Yale AI Engineering Research Grant from Yale Office of the Provost, and an LEAP-U Sponsored Research from Samsung Research America. Additionally, this research has greatly benefited from the discussions and research talks held at the IMS-NTU Joint Workshop on Applied Geometry for Data Sciences.  
\end{acks}

\newpage
\bibliographystyle{ACM-Reference-Format}
\bibliography{main}


\begin{thebibliography}{126}


\ifx \showCODEN    \undefined \def \showCODEN     #1{\unskip}     \fi
\ifx \showDOI      \undefined \def \showDOI       #1{#1}\fi
\ifx \showISBNx    \undefined \def \showISBNx     #1{\unskip}     \fi
\ifx \showISBNxiii \undefined \def \showISBNxiii  #1{\unskip}     \fi
\ifx \showISSN     \undefined \def \showISSN      #1{\unskip}     \fi
\ifx \showLCCN     \undefined \def \showLCCN      #1{\unskip}     \fi
\ifx \shownote     \undefined \def \shownote      #1{#1}          \fi
\ifx \showarticletitle \undefined \def \showarticletitle #1{#1}   \fi
\ifx \showURL      \undefined \def \showURL       {\relax}        \fi
\providecommand\bibfield[2]{#2}
\providecommand\bibinfo[2]{#2}
\providecommand\natexlab[1]{#1}
\providecommand\showeprint[2][]{arXiv:#2}

\bibitem[Alanis-Lobato et~al\mbox{.}(2018)]%
        {alanis2018latent}
\bibfield{author}{\bibinfo{person}{Gregorio Alanis-Lobato}, \bibinfo{person}{Pablo Mier}, {and} \bibinfo{person}{Miguel Andrade-Navarro}.} \bibinfo{year}{2018}\natexlab{}.
\newblock \showarticletitle{The latent geometry of the human protein interaction network}.
\newblock \bibinfo{journal}{\emph{Bioinformatics}} \bibinfo{volume}{34}, \bibinfo{number}{16} (\bibinfo{year}{2018}), \bibinfo{pages}{2826--2834}.
\newblock


\bibitem[Awais et~al\mbox{.}(2025)]%
        {awais2025foundation}
\bibfield{author}{\bibinfo{person}{Muhammad Awais}, \bibinfo{person}{Muzammal Naseer}, \bibinfo{person}{Salman Khan}, \bibinfo{person}{Rao~Muhammad Anwer}, \bibinfo{person}{Hisham Cholakkal}, \bibinfo{person}{Mubarak Shah}, \bibinfo{person}{Ming-Hsuan Yang}, {and} \bibinfo{person}{Fahad~Shahbaz Khan}.} \bibinfo{year}{2025}\natexlab{}.
\newblock \showarticletitle{Foundation Models Defining a New Era in Vision: a Survey and Outlook}.
\newblock \bibinfo{journal}{\emph{TPAMI}} (\bibinfo{year}{2025}).
\newblock


\bibitem[Ba et~al\mbox{.}(2016)]%
        {ba2016layer}
\bibfield{author}{\bibinfo{person}{Jimmy~Lei Ba}, \bibinfo{person}{Jamie~Ryan Kiros}, {and} \bibinfo{person}{Geoffrey~E. Hinton}.} \bibinfo{year}{2016}\natexlab{}.
\newblock \showarticletitle{Layer Normalization}.
\newblock \bibinfo{journal}{\emph{arXiv arXiv:1607.06450}} (\bibinfo{year}{2016}).
\newblock


\bibitem[Bai et~al\mbox{.}(2021)]%
        {bai2021modeling}
\bibfield{author}{\bibinfo{person}{Yushi Bai}, \bibinfo{person}{Zhitao Ying}, \bibinfo{person}{Hongyu Ren}, {and} \bibinfo{person}{Jure Leskovec}.} \bibinfo{year}{2021}\natexlab{}.
\newblock \showarticletitle{Modeling heterogeneous hierarchies with relation-specific hyperbolic cones}.
\newblock \bibinfo{journal}{\emph{NeurIPS}}  \bibinfo{volume}{34} (\bibinfo{year}{2021}), \bibinfo{pages}{12316--12327}.
\newblock


\bibitem[Barbero et~al\mbox{.}(2025)]%
        {barbero2025rope}
\bibfield{author}{\bibinfo{person}{Federico Barbero}, \bibinfo{person}{Alex Vitvitskyi}, \bibinfo{person}{Christos Perivolaropoulos}, \bibinfo{person}{Razvan Pascanu}, {and} \bibinfo{person}{Petar Veličković}.} \bibinfo{year}{2025}\natexlab{}.
\newblock \showarticletitle{Round and Round We Go! What makes Rotary Positional Encodings useful?}
\newblock \bibinfo{journal}{\emph{arXiv:2410.06205}} (\bibinfo{year}{2025}).
\newblock


\bibitem[Bdeir et~al\mbox{.}(2024)]%
        {Bdeir2024fully}
\bibfield{author}{\bibinfo{person}{Ahmad Bdeir}, \bibinfo{person}{Kristian Schwethelm}, {and} \bibinfo{person}{Niels Landwehr}.} \bibinfo{year}{2024}\natexlab{}.
\newblock \showarticletitle{Fully Hyperbolic Convolutional Neural Networks for Computer Vision}. In \bibinfo{booktitle}{\emph{ICLR}}.
\newblock


\bibitem[Bhasker et~al\mbox{.}(2024)]%
        {bhasker2024contrastive}
\bibfield{author}{\bibinfo{person}{Nithya Bhasker}, \bibinfo{person}{Hattie Chung}, \bibinfo{person}{Louis Boucherie}, \bibinfo{person}{Vladislav Kim}, \bibinfo{person}{Stefanie Speidel}, {and} \bibinfo{person}{Melanie Weber}.} \bibinfo{year}{2024}\natexlab{}.
\newblock \showarticletitle{Contrastive poincar{\'e} maps for single-cell data analysis}. In \bibinfo{booktitle}{\emph{ICLR 2024 Workshop on Machine Learning for Genomics Explorations}}.
\newblock


\bibitem[Bommasani and et~al.(2021)]%
        {bommasani2021opportunities}
\bibfield{author}{\bibinfo{person}{Rishi Bommasani} {and} \bibinfo{person}{et al.}} \bibinfo{year}{2021}\natexlab{}.
\newblock \showarticletitle{On the Opportunities and Risks of Foundation Models}.
\newblock \bibinfo{journal}{\emph{arXiv preprint}}  \bibinfo{volume}{arXiv:2108.07258} (\bibinfo{year}{2021}).
\newblock


\bibitem[Bose et~al\mbox{.}(2020)]%
        {bose2020latent}
\bibfield{author}{\bibinfo{person}{Joey Bose}, \bibinfo{person}{Ariella Smofsky}, \bibinfo{person}{Renjie Liao}, \bibinfo{person}{Prakash Panangaden}, {and} \bibinfo{person}{Will Hamilton}.} \bibinfo{year}{2020}\natexlab{}.
\newblock \showarticletitle{Latent variable modelling with hyperbolic normalizing flows}. In \bibinfo{booktitle}{\emph{ICML}}. PMLR, \bibinfo{pages}{1045--1055}.
\newblock


\bibitem[Brown et~al\mbox{.}(2020)]%
        {brown2020language}
\bibfield{author}{\bibinfo{person}{Tom Brown}, \bibinfo{person}{Benjamin Mann}, \bibinfo{person}{Nick Ryder}, \bibinfo{person}{Melanie Subbiah}, \bibinfo{person}{Jared~D Kaplan}, \bibinfo{person}{Prafulla Dhariwal}, \bibinfo{person}{Arvind Neelakantan}, \bibinfo{person}{Pranav Shyam}, \bibinfo{person}{Girish Sastry}, \bibinfo{person}{Amanda Askell}, {et~al\mbox{.}}} \bibinfo{year}{2020}\natexlab{}.
\newblock \showarticletitle{Language models are few-shot learners}.
\newblock \bibinfo{journal}{\emph{NeurIPS}}  \bibinfo{volume}{33} (\bibinfo{year}{2020}), \bibinfo{pages}{1877--1901}.
\newblock


\bibitem[Caffagni et~al\mbox{.}(2024)]%
        {caffagni2024revolution}
\bibfield{author}{\bibinfo{person}{Davide Caffagni}, \bibinfo{person}{Federico Cocchi}, \bibinfo{person}{Luca Barsellotti}, \bibinfo{person}{Nicholas Moratelli}, \bibinfo{person}{Sara Sarto}, \bibinfo{person}{Lorenzo Baraldi}, \bibinfo{person}{Marcella Cornia}, {and} \bibinfo{person}{Rita Cucchiara}.} \bibinfo{year}{2024}\natexlab{}.
\newblock \showarticletitle{The revolution of multimodal large language models: a survey}.
\newblock \bibinfo{journal}{\emph{arXiv:2402.12451}} (\bibinfo{year}{2024}).
\newblock


\bibitem[Cao et~al\mbox{.}(2024)]%
        {cao2024survey}
\bibfield{author}{\bibinfo{person}{Hanqun Cao}, \bibinfo{person}{Cheng Tan}, \bibinfo{person}{Zhangyang Gao}, \bibinfo{person}{Yilun Xu}, \bibinfo{person}{Guangyong Chen}, \bibinfo{person}{Pheng-Ann Heng}, {and} \bibinfo{person}{Stan~Z Li}.} \bibinfo{year}{2024}\natexlab{}.
\newblock \showarticletitle{A survey on generative diffusion models}.
\newblock \bibinfo{journal}{\emph{TKDE}} (\bibinfo{year}{2024}).
\newblock


\bibitem[Chami et~al\mbox{.}(2019)]%
        {hgcn2019}
\bibfield{author}{\bibinfo{person}{Ines Chami}, \bibinfo{person}{Zhitao Ying}, \bibinfo{person}{Christopher R{\'e}}, {and} \bibinfo{person}{Jure Leskovec}.} \bibinfo{year}{2019}\natexlab{}.
\newblock \showarticletitle{Hyperbolic graph convolutional neural networks}. In \bibinfo{booktitle}{\emph{NeurIPS}}. \bibinfo{pages}{4868--4879}.
\newblock


\bibitem[Chen et~al\mbox{.}(2022a)]%
        {chen2022interpretable}
\bibfield{author}{\bibinfo{person}{Jiayang Chen}, \bibinfo{person}{Zhihang Hu}, \bibinfo{person}{Siqi Sun}, \bibinfo{person}{Qingxiong Tan}, \bibinfo{person}{Yixuan Wang}, \bibinfo{person}{Qinze Yu}, \bibinfo{person}{Licheng Zong}, \bibinfo{person}{Liang Hong}, \bibinfo{person}{Jin Xiao}, \bibinfo{person}{Tao Shen}, {et~al\mbox{.}}} \bibinfo{year}{2022}\natexlab{a}.
\newblock \showarticletitle{Interpretable RNA foundation model from unannotated data for highly accurate RNA structure and function predictions}.
\newblock \bibinfo{journal}{\emph{arXiv:2204.00300}} (\bibinfo{year}{2022}).
\newblock


\bibitem[Chen et~al\mbox{.}(2020b)]%
        {chen2020simple}
\bibfield{author}{\bibinfo{person}{Ming Chen}, \bibinfo{person}{Zhewei Wei}, \bibinfo{person}{Zengfeng Huang}, \bibinfo{person}{Bolin Ding}, {and} \bibinfo{person}{Yaliang Li}.} \bibinfo{year}{2020}\natexlab{b}.
\newblock \showarticletitle{Simple and deep graph convolutional networks}. In \bibinfo{booktitle}{\emph{ICML}}. PMLR, \bibinfo{pages}{1725--1735}.
\newblock


\bibitem[Chen and Lipman(2024)]%
        {chen2024flow}
\bibfield{author}{\bibinfo{person}{Ricky T.~Q. Chen} {and} \bibinfo{person}{Yaron Lipman}.} \bibinfo{year}{2024}\natexlab{}.
\newblock \showarticletitle{Flow Matching on General Geometries}. In \bibinfo{booktitle}{\emph{ICLR}}.
\newblock


\bibitem[Chen et~al\mbox{.}(2024)]%
        {chen2024hyperbolic}
\bibfield{author}{\bibinfo{person}{Weize Chen}, \bibinfo{person}{Xu Han}, \bibinfo{person}{Yankai Lin}, \bibinfo{person}{Kaichen He}, \bibinfo{person}{Ruobing Xie}, \bibinfo{person}{Jie Zhou}, {and} \bibinfo{person}{Zhiyuan Liu}.} \bibinfo{year}{2024}\natexlab{}.
\newblock \showarticletitle{Hyperbolic Pre-Trained Language Model}.
\newblock \bibinfo{journal}{\emph{IEEE TASLP}}  \bibinfo{volume}{32} (\bibinfo{year}{2024}).
\newblock


\bibitem[Chen et~al\mbox{.}(2021)]%
        {chen2021fully}
\bibfield{author}{\bibinfo{person}{Weize Chen}, \bibinfo{person}{Xu Han}, \bibinfo{person}{Yankai Lin}, \bibinfo{person}{Hexu Zhao}, \bibinfo{person}{Zhiyuan Liu}, \bibinfo{person}{Peng Li}, \bibinfo{person}{Maosong Sun}, {and} \bibinfo{person}{Jie Zhou}.} \bibinfo{year}{2021}\natexlab{}.
\newblock \showarticletitle{Fully Hyperbolic Neural Networks}.
\newblock \bibinfo{journal}{\emph{arXiv:2105.14686}} (\bibinfo{year}{2021}).
\newblock


\bibitem[Chen et~al\mbox{.}(2020a)]%
        {chen2020improved}
\bibfield{author}{\bibinfo{person}{Xinlei Chen}, \bibinfo{person}{Haoqi Fan}, \bibinfo{person}{Ross Girshick}, {and} \bibinfo{person}{Kaiming He}.} \bibinfo{year}{2020}\natexlab{a}.
\newblock \showarticletitle{Improved baselines with momentum contrastive learning}.
\newblock \bibinfo{journal}{\emph{arXiv preprint arXiv:2003.04297,}} (\bibinfo{year}{2020}).
\newblock


\bibitem[Chen et~al\mbox{.}(2022b)]%
        {chen2021modeling}
\bibfield{author}{\bibinfo{person}{Yankai Chen}, \bibinfo{person}{Menglin Yang}, \bibinfo{person}{Yingxue Zhang}, \bibinfo{person}{Mengchen Zhao}, \bibinfo{person}{Ziqiao Meng}, \bibinfo{person}{Jianye Hao}, {and} \bibinfo{person}{Irwin King}.} \bibinfo{year}{2022}\natexlab{b}.
\newblock \showarticletitle{Modeling Scale-free Graphs for Knowledge-aware Recommendation}.
\newblock \bibinfo{journal}{\emph{WSDM}} (\bibinfo{year}{2022}).
\newblock


\bibitem[Cheng et~al\mbox{.}(2024)]%
        {cheng2024dated}
\bibfield{author}{\bibinfo{person}{Jeffrey Cheng}, \bibinfo{person}{Marc Marone}, \bibinfo{person}{Orion Weller}, \bibinfo{person}{Dawn Lawrie}, \bibinfo{person}{Daniel Khashabi}, {and} \bibinfo{person}{Benjamin Van~Durme}.} \bibinfo{year}{2024}\natexlab{}.
\newblock \showarticletitle{Dated data: Tracing knowledge cutoffs in large language models}.
\newblock \bibinfo{journal}{\emph{arXiv:2403.12958}} (\bibinfo{year}{2024}).
\newblock


\bibitem[Coenen et~al\mbox{.}(2019)]%
        {coenen2019visualizing}
\bibfield{author}{\bibinfo{person}{Andy Coenen}, \bibinfo{person}{Emily Reif}, \bibinfo{person}{Ann Yuan}, \bibinfo{person}{Been Kim}, \bibinfo{person}{Adam Pearce}, \bibinfo{person}{Fernanda Viégas}, {and} \bibinfo{person}{Martin Wattenberg}.} \bibinfo{year}{2019}\natexlab{}.
\newblock \showarticletitle{Visualizing and Measuring the Geometry of BERT}.
\newblock \bibinfo{journal}{\emph{NuerIPS}} (\bibinfo{year}{2019}).
\newblock


\bibitem[Croitoru et~al\mbox{.}(2023)]%
        {croitoru2023diffusion}
\bibfield{author}{\bibinfo{person}{Florinel-Alin Croitoru}, \bibinfo{person}{Vlad Hondru}, \bibinfo{person}{Radu~Tudor Ionescu}, {and} \bibinfo{person}{Mubarak Shah}.} \bibinfo{year}{2023}\natexlab{}.
\newblock \showarticletitle{Diffusion models in vision: A survey}.
\newblock \bibinfo{journal}{\emph{TPAMI}} (\bibinfo{year}{2023}).
\newblock


\bibitem[Cui et~al\mbox{.}(2024)]%
        {cui2024scgpt}
\bibfield{author}{\bibinfo{person}{Haotian Cui}, \bibinfo{person}{Chloe Wang}, \bibinfo{person}{Hassaan Maan}, \bibinfo{person}{Kuan Pang}, \bibinfo{person}{Fengning Luo}, \bibinfo{person}{Nan Duan}, {and} \bibinfo{person}{Bo Wang}.} \bibinfo{year}{2024}\natexlab{}.
\newblock \showarticletitle{scGPT: toward building a foundation model for single-cell multi-omics using generative AI}.
\newblock \bibinfo{journal}{\emph{Nature Methods}} \bibinfo{volume}{21}, \bibinfo{number}{8} (\bibinfo{year}{2024}), \bibinfo{pages}{1470--1480}.
\newblock


\bibitem[Dai et~al\mbox{.}(2021)]%
        {dai2021H2H}
\bibfield{author}{\bibinfo{person}{Jindou Dai}, \bibinfo{person}{Yuwei Wu}, \bibinfo{person}{Zhi Gao}, {and} \bibinfo{person}{Yunde Jia}.} \bibinfo{year}{2021}\natexlab{}.
\newblock \showarticletitle{A Hyperbolic-to-Hyperbolic Graph Convolutional Network}.
\newblock \bibinfo{journal}{\emph{arXiv:2104.06942}} (\bibinfo{year}{2021}), \bibinfo{pages}{154--163}.
\newblock


\bibitem[Dao et~al\mbox{.}(2022)]%
        {dao2022flashattention}
\bibfield{author}{\bibinfo{person}{Tri Dao}, \bibinfo{person}{Daniel~Y. Fu}, \bibinfo{person}{Stefano Ermon}, \bibinfo{person}{Atri Rudra}, {and} \bibinfo{person}{Christopher R{\'e}}.} \bibinfo{year}{2022}\natexlab{}.
\newblock \showarticletitle{Flash{A}ttention: Fast and Memory-Efficient Exact Attention with {IO}-Awareness}. In \bibinfo{booktitle}{\emph{NeurIPS}}.
\newblock


\bibitem[De~Bortoli et~al\mbox{.}(2022)]%
        {debortoli2022riemannian}
\bibfield{author}{\bibinfo{person}{Valentin De~Bortoli}, \bibinfo{person}{Emile Mathieu}, \bibinfo{person}{Michael Hutchinson}, \bibinfo{person}{James Thornton}, \bibinfo{person}{Yee~Whye Teh}, {and} \bibinfo{person}{Arnaud Doucet}.} \bibinfo{year}{2022}\natexlab{}.
\newblock \showarticletitle{Riemannian Score-Based Generative Modelling}. In \bibinfo{booktitle}{\emph{NeurIPS}}.
\newblock


\bibitem[DeepSeek-AI(2024)]%
        {deepseekv3}
\bibfield{author}{\bibinfo{person}{DeepSeek-AI}.} \bibinfo{year}{2024}\natexlab{}.
\newblock \showarticletitle{DeepSeek-V3 Technical Report}.
\newblock \bibinfo{journal}{\emph{arXiv:2412.19437}} (\bibinfo{year}{2024}).
\newblock


\bibitem[Desai et~al\mbox{.}(2023)]%
        {desai2023hyperbolic}
\bibfield{author}{\bibinfo{person}{Karan Desai}, \bibinfo{person}{Maximilian Nickel}, \bibinfo{person}{Tanmay Rajpurohit}, \bibinfo{person}{Justin Johnson}, {and} \bibinfo{person}{Shanmukha~Ramakrishna Vedantam}.} \bibinfo{year}{2023}\natexlab{}.
\newblock \showarticletitle{Hyperbolic image-text representations}. In \bibinfo{booktitle}{\emph{ICML}}. PMLR, \bibinfo{pages}{7694--7731}.
\newblock


\bibitem[Devlin et~al\mbox{.}(2018)]%
        {devlin2018bert}
\bibfield{author}{\bibinfo{person}{Jacob Devlin}, \bibinfo{person}{Ming-Wei Chang}, \bibinfo{person}{Kenton Lee}, {and} \bibinfo{person}{Kristina Toutanova}.} \bibinfo{year}{2018}\natexlab{}.
\newblock \showarticletitle{Bert: Pre-training of deep bidirectional transformers for language understanding}.
\newblock \bibinfo{journal}{\emph{arXiv:1810.04805}} (\bibinfo{year}{2018}).
\newblock


\bibitem[Ding and Regev(2021)]%
        {ding2021deep}
\bibfield{author}{\bibinfo{person}{Jiarui Ding} {and} \bibinfo{person}{Aviv Regev}.} \bibinfo{year}{2021}\natexlab{}.
\newblock \showarticletitle{Deep generative model embedding of single-cell RNA-Seq profiles on hyperspheres and hyperbolic spaces}.
\newblock \bibinfo{journal}{\emph{Nature communications}} \bibinfo{volume}{12}, \bibinfo{number}{1} (\bibinfo{year}{2021}), \bibinfo{pages}{2554}.
\newblock


\bibitem[Dosovitskiy et~al\mbox{.}(2020)]%
        {dosovitskiy2020image}
\bibfield{author}{\bibinfo{person}{Alexey Dosovitskiy}, \bibinfo{person}{Lucas Beyer}, \bibinfo{person}{Alexander Kolesnikov}, \bibinfo{person}{Dirk Weissenborn}, \bibinfo{person}{Xiaohua Zhai}, \bibinfo{person}{Thomas Unterthiner}, \bibinfo{person}{Mostafa Dehghani}, \bibinfo{person}{Matthias Minderer}, \bibinfo{person}{Georg Heigold}, \bibinfo{person}{Sylvain Gelly}, {et~al\mbox{.}}} \bibinfo{year}{2020}\natexlab{}.
\newblock \showarticletitle{An image is worth 16x16 words: Transformers for image recognition at scale}.
\newblock \bibinfo{journal}{\emph{arXiv:2010.11929}} (\bibinfo{year}{2020}).
\newblock


\bibitem[Edge et~al\mbox{.}(2024)]%
        {edge2024graphrag}
\bibfield{author}{\bibinfo{person}{Darren Edge}, \bibinfo{person}{Ha Trinh}, \bibinfo{person}{Newman Cheng}, \bibinfo{person}{Joshua Bradley}, \bibinfo{person}{Alex Chao}, \bibinfo{person}{Apurva Mody}, {and} \bibinfo{person}{Steven Truitt}.} \bibinfo{year}{2024}\natexlab{}.
\newblock \showarticletitle{From Local to Global: A Graph RAG Approach to Query-Focused Summarization}.
\newblock \bibinfo{journal}{\emph{arXiv:2404.16130}} (\bibinfo{year}{2024}).
\newblock


\bibitem[Ermolov et~al\mbox{.}(2022)]%
        {ermolov2022hyperbolic}
\bibfield{author}{\bibinfo{person}{Aleksandr Ermolov}, \bibinfo{person}{Leyla Mirvakhabova}, \bibinfo{person}{Valentin Khrulkov}, \bibinfo{person}{Nicu Sebe}, {and} \bibinfo{person}{Ivan Oseledets}.} \bibinfo{year}{2022}\natexlab{}.
\newblock \showarticletitle{Hyperbolic vision transformers: Combining improvements in metric learning}. In \bibinfo{booktitle}{\emph{CVPR}}. \bibinfo{pages}{7409--7419}.
\newblock


\bibitem[Fein-Ashley et~al\mbox{.}(2024)]%
        {fein-ashley2024hvt}
\bibfield{author}{\bibinfo{person}{Jacob Fein-Ashley}, \bibinfo{person}{Ethan Feng}, {and} \bibinfo{person}{Minh Pham}.} \bibinfo{year}{2024}\natexlab{}.
\newblock \showarticletitle{HVT: A Comprehensive Vision Framework for Learning in Non-Euclidean Space}.
\newblock \bibinfo{journal}{\emph{arXiv:2409.16897}} (\bibinfo{year}{2024}).
\newblock


\bibitem[Fu et~al\mbox{.}(2024)]%
        {fu2024hyperbolic}
\bibfield{author}{\bibinfo{person}{Xingcheng Fu}, \bibinfo{person}{Yisen Gao}, \bibinfo{person}{Yuecen Wei}, \bibinfo{person}{Qingyun Sun}, \bibinfo{person}{Hao Peng}, \bibinfo{person}{Jianxin Li}, {and} \bibinfo{person}{Xianxian Li}.} \bibinfo{year}{2024}\natexlab{}.
\newblock \showarticletitle{Hyperbolic Geometric Latent Diffusion Model for Graph Generation}.
\newblock \bibinfo{journal}{\emph{ICML}} (\bibinfo{year}{2024}).
\newblock


\bibitem[Ganea et~al\mbox{.}(2018a)]%
        {ganea2018hyperbolic}
\bibfield{author}{\bibinfo{person}{Octavian Ganea}, \bibinfo{person}{Gary B{\'e}cigneul}, {and} \bibinfo{person}{Thomas Hofmann}.} \bibinfo{year}{2018}\natexlab{a}.
\newblock \showarticletitle{Hyperbolic entailment cones for learning hierarchical embeddings}. In \bibinfo{booktitle}{\emph{ICML}}. PMLR, \bibinfo{pages}{1646--1655}.
\newblock


\bibitem[Ganea et~al\mbox{.}(2018b)]%
        {HNN}
\bibfield{author}{\bibinfo{person}{Octavian Ganea}, \bibinfo{person}{Gary B{\'e}cigneul}, {and} \bibinfo{person}{Thomas Hofmann}.} \bibinfo{year}{2018}\natexlab{b}.
\newblock \showarticletitle{Hyperbolic neural networks}. In \bibinfo{booktitle}{\emph{NeurIPS}}. \bibinfo{pages}{5345--5355}.
\newblock


\bibitem[Gao et~al\mbox{.}(2023)]%
        {gao2023retrieval}
\bibfield{author}{\bibinfo{person}{Yunfan Gao}, \bibinfo{person}{Yun Xiong}, \bibinfo{person}{Xinyu Gao}, \bibinfo{person}{Kangxiang Jia}, \bibinfo{person}{Jinliu Pan}, \bibinfo{person}{Yuxi Bi}, \bibinfo{person}{Yi Dai}, \bibinfo{person}{Jiawei Sun}, \bibinfo{person}{Haofen Wang}, {and} \bibinfo{person}{Haofen Wang}.} \bibinfo{year}{2023}\natexlab{}.
\newblock \showarticletitle{Retrieval-augmented generation for large language models: A survey}.
\newblock \bibinfo{journal}{\emph{arXiv:2312.10997}}  \bibinfo{volume}{2} (\bibinfo{year}{2023}).
\newblock


\bibitem[Ge et~al\mbox{.}(2023)]%
        {ge2023hyperbolic}
\bibfield{author}{\bibinfo{person}{Songwei Ge}, \bibinfo{person}{Shlok Mishra}, \bibinfo{person}{Simon Kornblith}, \bibinfo{person}{Chun-Liang Li}, {and} \bibinfo{person}{David Jacobs}.} \bibinfo{year}{2023}\natexlab{}.
\newblock \showarticletitle{Hyperbolic contrastive learning for visual representations beyond objects}. In \bibinfo{booktitle}{\emph{CVPR}}. \bibinfo{pages}{6840--6849}.
\newblock


\bibitem[Ge et~al\mbox{.}(2022)]%
        {Ge2022HyperbolicCL}
\bibfield{author}{\bibinfo{person}{Songwei Ge}, \bibinfo{person}{Shlok~Kumar Mishra}, \bibinfo{person}{Simon Kornblith}, \bibinfo{person}{Chun-Liang Li}, {and} \bibinfo{person}{David Jacobs}.} \bibinfo{year}{2022}\natexlab{}.
\newblock \showarticletitle{Hyperbolic Contrastive Learning for Visual Representations beyond Objects}.
\newblock \bibinfo{journal}{\emph{ArXiv}}  \bibinfo{volume}{abs/2212.00653} (\bibinfo{year}{2022}).
\newblock


\bibitem[Gower(1985)]%
        {gower1985properties}
\bibfield{author}{\bibinfo{person}{John~Clifford Gower}.} \bibinfo{year}{1985}\natexlab{}.
\newblock \showarticletitle{Properties of Euclidean and non-Euclidean distance matrices}.
\newblock \bibinfo{journal}{\emph{Linear algebra and its applications}}  \bibinfo{volume}{67} (\bibinfo{year}{1985}), \bibinfo{pages}{81--97}.
\newblock


\bibitem[Grattafiori et~al\mbox{.}(2024)]%
        {dubey2024llama3}
\bibfield{author}{\bibinfo{person}{Aaron Grattafiori}, \bibinfo{person}{Abhimanyu Dubey}, \bibinfo{person}{Abhinav Jauhri}, \bibinfo{person}{Abhinav Pandey}, \bibinfo{person}{Abhishek Kadian}, \bibinfo{person}{Ahmad Al-Dahle}, \bibinfo{person}{Aiesha Letman}, \bibinfo{person}{Akhil Mathur}, \bibinfo{person}{Alan Schelten}, \bibinfo{person}{Amy Yang}, \bibinfo{person}{Angela Fan}, {et~al\mbox{.}}} \bibinfo{year}{2024}\natexlab{}.
\newblock \showarticletitle{The Llama 3 Herd of Models}.
\newblock \bibinfo{journal}{\emph{arXiv:2407.21783}} (\bibinfo{year}{2024}).
\newblock


\bibitem[Gulcehre et~al\mbox{.}(2019)]%
        {gulcehre2019hyperbolicAT}
\bibfield{author}{\bibinfo{person}{Caglar Gulcehre}, \bibinfo{person}{Misha Denil}, \bibinfo{person}{Mateusz Malinowski}, \bibinfo{person}{Ali Razavi}, \bibinfo{person}{Razvan Pascanu}, \bibinfo{person}{Karl~Moritz Hermann}, \bibinfo{person}{Peter Battaglia}, \bibinfo{person}{Victor Bapst}, \bibinfo{person}{David Raposo}, \bibinfo{person}{Adam Santoro}, {et~al\mbox{.}}} \bibinfo{year}{2019}\natexlab{}.
\newblock \showarticletitle{Hyperbolic attention networks}. In \bibinfo{booktitle}{\emph{ICLR}}.
\newblock


\bibitem[Guo et~al\mbox{.}(2025)]%
        {guo2025graphmore}
\bibfield{author}{\bibinfo{person}{Zihao Guo}, \bibinfo{person}{Qingyun Sun}, \bibinfo{person}{Haonan Yuan}, \bibinfo{person}{Xingcheng Fu}, \bibinfo{person}{Min Zhou}, \bibinfo{person}{Yisen Gao}, {and} \bibinfo{person}{Jianxin Li}.} \bibinfo{year}{2025}\natexlab{}.
\newblock \showarticletitle{GraphMoRE: Mitigating Topological Heterogeneity via Mixture of Riemannian Experts}. In \bibinfo{booktitle}{\emph{AAAI}}.
\newblock


\bibitem[He et~al\mbox{.}(2020)]%
        {he2020momentum}
\bibfield{author}{\bibinfo{person}{Kaiming He}, \bibinfo{person}{Haoqi Fan}, \bibinfo{person}{Yuxin Wu}, \bibinfo{person}{Saining Xie}, {and} \bibinfo{person}{Ross Girshick}.} \bibinfo{year}{2020}\natexlab{}.
\newblock \showarticletitle{Momentum contrast for unsupervised visual repre- sentation learning}. In \bibinfo{booktitle}{\emph{CVPR}}.
\newblock


\bibitem[He et~al\mbox{.}(2025a)]%
        {he2025helm}
\bibfield{author}{\bibinfo{person}{Neil He}, \bibinfo{person}{Rishabh Anand}, \bibinfo{person}{Hiren Madhu}, \bibinfo{person}{Ali Maatouk}, \bibinfo{person}{Smita Krishnaswamy}, \bibinfo{person}{Leandros Tassiulas}, \bibinfo{person}{Menglin Yang}, {and} \bibinfo{person}{Rex Ying}.} \bibinfo{year}{2025}\natexlab{a}.
\newblock \showarticletitle{HELM: Hyperbolic Large Language Models via Mixture-of-Curvature Experts}.
\newblock \bibinfo{journal}{\emph{arXiv preprint arXiv:2505.24722}} (\bibinfo{year}{2025}).
\newblock


\bibitem[He et~al\mbox{.}(2025b)]%
        {he2025position}
\bibfield{author}{\bibinfo{person}{Neil He}, \bibinfo{person}{Jiahong Liu}, \bibinfo{person}{Buze Zhang}, \bibinfo{person}{Ngoc Bui}, \bibinfo{person}{Ali Maatouk}, \bibinfo{person}{Menglin Yang}, \bibinfo{person}{Irwin King}, \bibinfo{person}{Melanie Weber}, {and} \bibinfo{person}{Rex Ying}.} \bibinfo{year}{2025}\natexlab{b}.
\newblock \showarticletitle{Position: Beyond Euclidean--Foundation Models Should Embrace Non-Euclidean Geometries}.
\newblock \bibinfo{journal}{\emph{arXiv:2504.08896}} (\bibinfo{year}{2025}).
\newblock


\bibitem[He et~al\mbox{.}(2025c)]%
        {he2025hypercore}
\bibfield{author}{\bibinfo{person}{Neil He}, \bibinfo{person}{Menglin Yang}, {and} \bibinfo{person}{Rex Ying}.} \bibinfo{year}{2025}\natexlab{c}.
\newblock \showarticletitle{HyperCore: The Core Framework for Building Hyperbolic Foundation Models with Comprehensive Modules}.
\newblock \bibinfo{journal}{\emph{arXiv:2504.08912}} (\bibinfo{year}{2025}).
\newblock


\bibitem[He et~al\mbox{.}(2025d)]%
        {he2025lresnet}
\bibfield{author}{\bibinfo{person}{Neil He}, \bibinfo{person}{Menglin Yang}, {and} \bibinfo{person}{Rex Ying}.} \bibinfo{year}{2025}\natexlab{d}.
\newblock \showarticletitle{Lorentzian Residual Neural Networks}. In \bibinfo{booktitle}{\emph{KDD}}.
\newblock


\bibitem[Ho et~al\mbox{.}(2020)]%
        {ho2020denoising}
\bibfield{author}{\bibinfo{person}{Jonathan Ho}, \bibinfo{person}{Ajay Jain}, {and} \bibinfo{person}{Pieter Abbeel}.} \bibinfo{year}{2020}\natexlab{}.
\newblock \showarticletitle{Denoising Diffusion Probabilistic Models}. In \bibinfo{booktitle}{\emph{NeurIPS}}.
\newblock


\bibitem[Hoffmann and et~al.(2022)]%
        {hoffman2022training}
\bibfield{author}{\bibinfo{person}{Jordan Hoffmann} {and} \bibinfo{person}{et al.}} \bibinfo{year}{2022}\natexlab{}.
\newblock \showarticletitle{Training compute-optimal large language models}. In \bibinfo{booktitle}{\emph{NeurIPS}}. \bibinfo{address}{Red Hook, NY, USA}, Article \bibinfo{articleno}{2176}, \bibinfo{numpages}{15}~pages.
\newblock
\showISBNx{9781713871088}


\bibitem[Hu et~al\mbox{.}(2022)]%
        {hu2022lora}
\bibfield{author}{\bibinfo{person}{Edward~J Hu}, \bibinfo{person}{Yelong Shen}, \bibinfo{person}{Phillip Wallis}, \bibinfo{person}{Zeyuan Allen-Zhu}, \bibinfo{person}{Yuanzhi Li}, \bibinfo{person}{Shean Wang}, \bibinfo{person}{Lu Wang}, {and} \bibinfo{person}{Weizhu Chen}.} \bibinfo{year}{2022}\natexlab{}.
\newblock \showarticletitle{Lo{RA}: Low-Rank Adaptation of Large Language Models}. In \bibinfo{booktitle}{\emph{ICLR}}.
\newblock


\bibitem[Huang et~al\mbox{.}(2022)]%
        {huang2022riemannian}
\bibfield{author}{\bibinfo{person}{Chin-Wei Huang}, \bibinfo{person}{Milad Aghajohari}, \bibinfo{person}{Avishek~Joey Bose}, \bibinfo{person}{Prakash Panangaden}, {and} \bibinfo{person}{Aaron Courville}.} \bibinfo{year}{2022}\natexlab{}.
\newblock \showarticletitle{Riemannian Diffusion Models}. In \bibinfo{booktitle}{\emph{NeurIPS}}.
\newblock


\bibitem[Huang et~al\mbox{.}(2025)]%
        {huang2025survey}
\bibfield{author}{\bibinfo{person}{Lei Huang}, \bibinfo{person}{Weijiang Yu}, \bibinfo{person}{Weitao Ma}, \bibinfo{person}{Weihong Zhong}, \bibinfo{person}{Zhangyin Feng}, \bibinfo{person}{Haotian Wang}, \bibinfo{person}{Qianglong Chen}, \bibinfo{person}{Weihua Peng}, \bibinfo{person}{Xiaocheng Feng}, \bibinfo{person}{Bing Qin}, {et~al\mbox{.}}} \bibinfo{year}{2025}\natexlab{}.
\newblock \showarticletitle{A survey on hallucination in large language models: Principles, taxonomy, challenges, and open questions}.
\newblock \bibinfo{journal}{\emph{TOIS}} \bibinfo{volume}{43}, \bibinfo{number}{2} (\bibinfo{year}{2025}), \bibinfo{pages}{1--55}.
\newblock


\bibitem[Ioffe and Szegedy(2015)]%
        {ioffe2015batch}
\bibfield{author}{\bibinfo{person}{Sergey Ioffe} {and} \bibinfo{person}{Christian Szegedy}.} \bibinfo{year}{2015}\natexlab{}.
\newblock \showarticletitle{Batch Normalization: Accelerating Deep Network Training by Reducing Internal Covariate Shift}. In \bibinfo{booktitle}{\emph{ICML}}. \bibinfo{pages}{448--456}.
\newblock


\bibitem[Jo and Lee(2022)]%
        {jo2022score}
\bibfield{author}{\bibinfo{person}{Jaehyeong Jo} {and} \bibinfo{person}{Hwang Sung~Ju Lee, Seul~and}.} \bibinfo{year}{2022}\natexlab{}.
\newblock \showarticletitle{Score-based Generative Modeling of Graphs via the System of Stochastic Differential Equations}. In \bibinfo{booktitle}{\emph{ICML}}.
\newblock


\bibitem[Kapoor et~al\mbox{.}(2023)]%
        {kapoor2023societal}
\bibfield{author}{\bibinfo{person}{Sayash Kapoor}, \bibinfo{person}{Rishi Bommasani}, \bibinfo{person}{Kevin Klyman}, \bibinfo{person}{Shayne Longpre}, \bibinfo{person}{Ashwin Ramaswami}, \bibinfo{person}{Peter Cihon}, \bibinfo{person}{Aspen Hopkins}, \bibinfo{person}{Kevin Bankston}, \bibinfo{person}{Stella Biderman}, \bibinfo{person}{Miranda Bogen}, \bibinfo{person}{Rumman Chowdhury}, \bibinfo{person}{Alex Engler}, \bibinfo{person}{Peter Henderson}, \bibinfo{person}{Yacine Jernite}, \bibinfo{person}{Seth Lazar}, \bibinfo{person}{Stefano Maffulli}, \bibinfo{person}{Alondra Nelson}, \bibinfo{person}{Joelle Pineau}, \bibinfo{person}{Aviya Skowron}, \bibinfo{person}{Dawn Song}, \bibinfo{person}{Victor Storchan}, \bibinfo{person}{Daniel Zhang}, \bibinfo{person}{Daniel~E. Ho}, \bibinfo{person}{Percy Liang}, {and} \bibinfo{person}{Arvind Narayanan}.} \bibinfo{year}{2023}\natexlab{}.
\newblock \showarticletitle{On the Societal Impact of Open Foundation Models}.
\newblock \bibinfo{journal}{\emph{arXiv:2304.11082}} (\bibinfo{year}{2023}).
\newblock


\bibitem[Kennedy et~al\mbox{.}(2013)]%
        {kennedy2013hyperbolicity}
\bibfield{author}{\bibinfo{person}{W~Sean Kennedy}, \bibinfo{person}{Onuttom Narayan}, {and} \bibinfo{person}{Iraj Saniee}.} \bibinfo{year}{2013}\natexlab{}.
\newblock \showarticletitle{On the hyperbolicity of large-scale networks}.
\newblock \bibinfo{journal}{\emph{arXiv:1307.0031}} (\bibinfo{year}{2013}).
\newblock


\bibitem[Kiani et~al\mbox{.}(2024)]%
        {kiani2024hardness}
\bibfield{author}{\bibinfo{person}{Bobak Kiani}, \bibinfo{person}{Jason Wang}, {and} \bibinfo{person}{Melanie Weber}.} \bibinfo{year}{2024}\natexlab{}.
\newblock \showarticletitle{Hardness of Learning Neural Networks under the Manifold Hypothesis}. In \bibinfo{booktitle}{\emph{NeurIPS}}.
\newblock


\bibitem[Kim et~al\mbox{.}(2020)]%
        {kim2020adversarial}
\bibfield{author}{\bibinfo{person}{Minseon Kim}, \bibinfo{person}{Jihoon Tack}, {and} \bibinfo{person}{Sung~Ju Hwang}.} \bibinfo{year}{2020}\natexlab{}.
\newblock \showarticletitle{Adversarial Self-Supervised Contrastive Learning}. In \bibinfo{booktitle}{\emph{NeurIPS}}.
\newblock


\bibitem[Kingma and Welling(2019)]%
        {kingma2019introduction}
\bibfield{author}{\bibinfo{person}{Diederik~P. Kingma} {and} \bibinfo{person}{Max Welling}.} \bibinfo{year}{2019}\natexlab{}.
\newblock \bibinfo{booktitle}{\emph{An Introduction to Variational Autoencoders}}. \bibinfo{series}{Foundations and Trends in Machine Learning}, Vol.~\bibinfo{volume}{12}.
\newblock 307--392 pages.
\newblock


\bibitem[Kochurov et~al\mbox{.}(2020)]%
        {kochurov2020geoopt}
\bibfield{author}{\bibinfo{person}{Max Kochurov}, \bibinfo{person}{Rasul Karimov}, {and} \bibinfo{person}{Sergei Kozlukov}.} \bibinfo{year}{2020}\natexlab{}.
\newblock \showarticletitle{Geoopt: Riemannian Optimization in PyTorch}.
\newblock \bibinfo{journal}{\emph{arXiv:2005.02819}} (\bibinfo{year}{2020}).
\newblock


\bibitem[Krioukov et~al\mbox{.}(2010)]%
        {krioukov2010hyperbolic}
\bibfield{author}{\bibinfo{person}{Dmitri Krioukov}, \bibinfo{person}{Fragkiskos Papadopoulos}, \bibinfo{person}{Maksim Kitsak}, \bibinfo{person}{Amin Vahdat}, {and} \bibinfo{person}{Mari{\'a}n Bogun{\'a}}.} \bibinfo{year}{2010}\natexlab{}.
\newblock \showarticletitle{Hyperbolic geometry of complex networks}.
\newblock \bibinfo{journal}{\emph{Physical Review E—Statistical, Nonlinear, and Soft Matter Physics}} \bibinfo{volume}{82}, \bibinfo{number}{3} (\bibinfo{year}{2010}), \bibinfo{pages}{036106}.
\newblock


\bibitem[Law et~al\mbox{.}(2019)]%
        {law2019lorentzian}
\bibfield{author}{\bibinfo{person}{Marc Law}, \bibinfo{person}{Renjie Liao}, \bibinfo{person}{Jake Snell}, {and} \bibinfo{person}{Richard Zemel}.} \bibinfo{year}{2019}\natexlab{}.
\newblock \showarticletitle{Lorentzian distance learning for hyperbolic representations}. In \bibinfo{booktitle}{\emph{ICML}}. PMLR, \bibinfo{pages}{3672--3681}.
\newblock


\bibitem[Lazcano et~al\mbox{.}(2021)]%
        {lazcano2021hyperbolic}
\bibfield{author}{\bibinfo{person}{Diego Lazcano}, \bibinfo{person}{Nicolás Fredes}, {and} \bibinfo{person}{Werner Creixell}.} \bibinfo{year}{2021}\natexlab{}.
\newblock \showarticletitle{Hyperbolic Generative Adversarial Network}.
\newblock \bibinfo{journal}{\emph{IEEE Access}}  \bibinfo{volume}{9} (\bibinfo{year}{2021}), \bibinfo{pages}{96309--96320}.
\newblock


\bibitem[Le et~al\mbox{.}(2019)]%
        {le2019inferring}
\bibfield{author}{\bibinfo{person}{Matthew Le}, \bibinfo{person}{Stephen Roller}, \bibinfo{person}{Laetitia Papaxanthos}, \bibinfo{person}{Douwe Kiela}, {and} \bibinfo{person}{Maximilian Nickel}.} \bibinfo{year}{2019}\natexlab{}.
\newblock \showarticletitle{Inferring Concept Hierarchies from Text Corpora via Hyperbolic Embeddings}. In \bibinfo{booktitle}{\emph{ACL}}. \bibinfo{pages}{3231--3241}.
\newblock


\bibitem[Lee et~al\mbox{.}(2022)]%
        {lee2022convergence}
\bibfield{author}{\bibinfo{person}{Holden Lee}, \bibinfo{person}{Jianfeng Lu}, {and} \bibinfo{person}{Yixin Tan}.} \bibinfo{year}{2022}\natexlab{}.
\newblock \showarticletitle{Convergence for score-based generative modeling with polynomial complexity}. In \bibinfo{booktitle}{\emph{NeurIPS}}.
\newblock


\bibitem[Lewis et~al\mbox{.}(2020)]%
        {lewis2020retrieval}
\bibfield{author}{\bibinfo{person}{Patrick Lewis}, \bibinfo{person}{Ethan Perez}, \bibinfo{person}{Aleksandra Piktus}, \bibinfo{person}{Fabio Petroni}, \bibinfo{person}{Vladimir Karpukhin}, \bibinfo{person}{Naman Goyal}, \bibinfo{person}{Heinrich K{\"u}ttler}, \bibinfo{person}{Mike Lewis}, \bibinfo{person}{Wen-tau Yih}, \bibinfo{person}{Tim Rockt{\"a}schel}, {et~al\mbox{.}}} \bibinfo{year}{2020}\natexlab{}.
\newblock \showarticletitle{Retrieval-Augmented Generation for Knowledge-Intensive NLP Tasks}. In \bibinfo{booktitle}{\emph{NeurIPS}}. \bibinfo{pages}{9459--9774}.
\newblock


\bibitem[Li et~al\mbox{.}(2023a)]%
        {li2023blip2}
\bibfield{author}{\bibinfo{person}{Junnan Li}, \bibinfo{person}{Dongxu Li}, \bibinfo{person}{Silvio Savarese}, {and} \bibinfo{person}{Steven Hoi}.} \bibinfo{year}{2023}\natexlab{a}.
\newblock \showarticletitle{{BLIP}-2: Bootstrapping Language-Image Pre-training with Frozen Image Encoders and Large Language Models}. In \bibinfo{booktitle}{\emph{ICML}}.
\newblock


\bibitem[Li et~al\mbox{.}(2023b)]%
        {li2023euclidean}
\bibfield{author}{\bibinfo{person}{Lingxiao Li}, \bibinfo{person}{Yi Zhang}, {and} \bibinfo{person}{Shuhui Wang}.} \bibinfo{year}{2023}\natexlab{b}.
\newblock \showarticletitle{The Euclidean Space is Evil: Hyperbolic Attribute Editing for Few-shot Image Generation}. In \bibinfo{booktitle}{\emph{ICCV}}.
\newblock


\bibitem[Linial et~al\mbox{.}(1995)]%
        {linial1995geometry}
\bibfield{author}{\bibinfo{person}{Nathan Linial}, \bibinfo{person}{Eran London}, {and} \bibinfo{person}{Yuri Rabinovich}.} \bibinfo{year}{1995}\natexlab{}.
\newblock \showarticletitle{The geometry of graphs and some of its algorithmic applications}.
\newblock \bibinfo{journal}{\emph{Combinatorica}} \bibinfo{volume}{15}, \bibinfo{number}{2} (\bibinfo{year}{1995}), \bibinfo{pages}{215--245}.
\newblock


\bibitem[Lipman et~al\mbox{.}(2022)]%
        {lipman2022flow}
\bibfield{author}{\bibinfo{person}{Yaron Lipman}, \bibinfo{person}{Ricky T.~Q. Chen}, \bibinfo{person}{Heli Ben-Hamu}, \bibinfo{person}{Maximilian Nickel}, {and} \bibinfo{person}{Matt Le}.} \bibinfo{year}{2022}\natexlab{}.
\newblock \showarticletitle{Flow Matching for Generative Modeling}.
\newblock \bibinfo{journal}{\emph{arXiv:2210.02747}} (\bibinfo{year}{2022}).
\newblock


\bibitem[Liu et~al\mbox{.}(2022)]%
        {liu2022enhancing}
\bibfield{author}{\bibinfo{person}{Jiahong Liu}, \bibinfo{person}{Menglin Yang}, \bibinfo{person}{Min Zhou}, \bibinfo{person}{Shanshan Feng}, {and} \bibinfo{person}{Philippe Fournier-Viger}.} \bibinfo{year}{2022}\natexlab{}.
\newblock \showarticletitle{Enhancing Hyperbolic Graph Embeddings via Contrastive Learning}. In \bibinfo{booktitle}{\emph{NeurIPS 2nd SSL workshop}}.
\newblock


\bibitem[Liu et~al\mbox{.}(2019)]%
        {liu2019HGNN}
\bibfield{author}{\bibinfo{person}{Qi Liu}, \bibinfo{person}{Maximilian Nickel}, {and} \bibinfo{person}{Douwe Kiela}.} \bibinfo{year}{2019}\natexlab{}.
\newblock \showarticletitle{Hyperbolic graph neural networks}. In \bibinfo{booktitle}{\emph{NeurIPS}}. \bibinfo{pages}{8230--8241}.
\newblock


\bibitem[Lou et~al\mbox{.}(2020a)]%
        {lou2020differentiating}
\bibfield{author}{\bibinfo{person}{Aaron Lou}, \bibinfo{person}{Isay Katsman}, \bibinfo{person}{Qingxuan Jiang}, \bibinfo{person}{Serge Belongie}, \bibinfo{person}{Ser-Nam Lim}, {and} \bibinfo{person}{Christopher De~Sa}.} \bibinfo{year}{2020}\natexlab{a}.
\newblock \showarticletitle{Differentiating through the Fr$\backslash$'echet Mean}.
\newblock \bibinfo{journal}{\emph{arXiv:2003.00335}} (\bibinfo{year}{2020}).
\newblock


\bibitem[Lou et~al\mbox{.}(2020b)]%
        {lou2020Frechet}
\bibfield{author}{\bibinfo{person}{Aaron Lou}, \bibinfo{person}{Isay Katsman}, \bibinfo{person}{Qingxuan Jiang}, \bibinfo{person}{Serge Belongie}, \bibinfo{person}{Ser-Nam Lim}, {and} \bibinfo{person}{Christopher De~Sa}.} \bibinfo{year}{2020}\natexlab{b}.
\newblock \showarticletitle{Differentiating through the Fr{\'e}chet Mean}. In \bibinfo{booktitle}{\emph{ICML}}. \bibinfo{pages}{6393--6403}.
\newblock


\bibitem[Ma et~al\mbox{.}(2024)]%
        {ma2024harec}
\bibfield{author}{\bibinfo{person}{Qiyao Ma}, \bibinfo{person}{Menglin Yang}, \bibinfo{person}{Mingxuan Ju}, \bibinfo{person}{Tong Zhao}, \bibinfo{person}{Neil Shah}, {and} \bibinfo{person}{Rex Ying}.} \bibinfo{year}{2024}\natexlab{}.
\newblock \showarticletitle{{HARec}: Hyperbolic graph-llm alignment for exploration and exploitation in recommender systems}.
\newblock \bibinfo{journal}{\emph{arXiv:2411.13865}} (\bibinfo{year}{2024}).
\newblock


\bibitem[Mandica et~al\mbox{.}(2024)]%
        {mandica2024hyperbolic}
\bibfield{author}{\bibinfo{person}{Paolo Mandica}, \bibinfo{person}{Luca Franco}, \bibinfo{person}{Konstantinos Kallidromitis}, \bibinfo{person}{Suzanne Petryk}, {and} \bibinfo{person}{Fabio Galasso}.} \bibinfo{year}{2024}\natexlab{}.
\newblock \showarticletitle{Hyperbolic Learning with Multimodal Large Language Models}.
\newblock \bibinfo{journal}{\emph{arXiv:2408.05097}} (\bibinfo{year}{2024}).
\newblock


\bibitem[Mathieu et~al\mbox{.}(2019)]%
        {mathieu2019continuous}
\bibfield{author}{\bibinfo{person}{Emile Mathieu}, \bibinfo{person}{Charline Le~Lan}, \bibinfo{person}{Chris~J. Maddison}, \bibinfo{person}{Ryota Tomioka}, {and} \bibinfo{person}{Yee~Whye Teh}.} \bibinfo{year}{2019}\natexlab{}.
\newblock \showarticletitle{Continuous Hierarchical Representations with Poincaré Variational Auto-Encoders}. In \bibinfo{booktitle}{\emph{NeurIPS}}.
\newblock


\bibitem[Mettes et~al\mbox{.}(2024)]%
        {mettes2024hyperbolic}
\bibfield{author}{\bibinfo{person}{Pascal Mettes}, \bibinfo{person}{Mina Ghadimi~Atigh}, \bibinfo{person}{Martin Keller-Ressel}, \bibinfo{person}{Jeffrey Gu}, {and} \bibinfo{person}{Serena Yeung}.} \bibinfo{year}{2024}\natexlab{}.
\newblock \showarticletitle{Hyperbolic deep learning in computer vision: A survey}.
\newblock \bibinfo{journal}{\emph{International Journal of Computer Vision}} (\bibinfo{year}{2024}), \bibinfo{pages}{1--25}.
\newblock


\bibitem[Miolane et~al\mbox{.}(2020)]%
        {geomstats}
\bibfield{author}{\bibinfo{person}{Nina Miolane}, \bibinfo{person}{Nicolas Guigui}, \bibinfo{person}{Alice~Le Brigant}, \bibinfo{person}{Johan Mathe}, \bibinfo{person}{Benjamin Hou}, \bibinfo{person}{Yann Thanwerdas}, \bibinfo{person}{Stefan Heyder}, \bibinfo{person}{Olivier Peltre}, \bibinfo{person}{Niklas Koep}, \bibinfo{person}{Hadi Zaatiti}, \bibinfo{person}{Hatem Hajri}, \bibinfo{person}{Yann Cabanes}, \bibinfo{person}{Thomas Gerald}, \bibinfo{person}{Paul Chauchat}, \bibinfo{person}{Christian Shewmake}, \bibinfo{person}{Daniel Brooks}, \bibinfo{person}{Bernhard Kainz}, \bibinfo{person}{Claire Donnat}, \bibinfo{person}{Susan Holmes}, {and} \bibinfo{person}{Xavier Pennec}.} \bibinfo{year}{2020}\natexlab{}.
\newblock \showarticletitle{Geomstats: A Python Package for Riemannian Geometry in Machine Learning}.
\newblock \bibinfo{journal}{\emph{JMLR}} \bibinfo{volume}{21}, \bibinfo{number}{223} (\bibinfo{year}{2020}), \bibinfo{pages}{1--9}.
\newblock


\bibitem[Myers et~al\mbox{.}(2024)]%
        {myers2024foundation}
\bibfield{author}{\bibinfo{person}{Devon Myers}, \bibinfo{person}{Rami Mohawesh}, \bibinfo{person}{Venkata~Ishwarya Chellaboina}, \bibinfo{person}{Anantha~Lakshmi Sathvik}, \bibinfo{person}{Praveen Venkatesh}, \bibinfo{person}{Yi-Hui Ho}, \bibinfo{person}{Hanna Henshaw}, \bibinfo{person}{Muna Alhawawreh}, \bibinfo{person}{David Berdik}, {and} \bibinfo{person}{Yaser Jararweh}.} \bibinfo{year}{2024}\natexlab{}.
\newblock \showarticletitle{Foundation and large language models: fundamentals, challenges, opportunities, and social impacts}.
\newblock \bibinfo{journal}{\emph{Cluster Computing}} \bibinfo{volume}{27}, \bibinfo{number}{1} (\bibinfo{year}{2024}), \bibinfo{pages}{1--26}.
\newblock


\bibitem[Nagano et~al\mbox{.}(2019)]%
        {nagano2019wrapped}
\bibfield{author}{\bibinfo{person}{Yoshihiro Nagano}, \bibinfo{person}{Shoichiro Yamaguchi}, \bibinfo{person}{Yasuhiro Fujita}, {and} \bibinfo{person}{Masanori Koyama}.} \bibinfo{year}{2019}\natexlab{}.
\newblock \showarticletitle{A wrapped normal distribution on hyperbolic space for gradient-based learning}. In \bibinfo{booktitle}{\emph{ICML}}. PMLR, \bibinfo{pages}{4693--4702}.
\newblock


\bibitem[Nickel and Kiela(2017)]%
        {nickel2017poincare}
\bibfield{author}{\bibinfo{person}{Maximillian Nickel} {and} \bibinfo{person}{Douwe Kiela}.} \bibinfo{year}{2017}\natexlab{}.
\newblock \showarticletitle{Poincar{\'e} embeddings for learning hierarchical representations}. In \bibinfo{booktitle}{\emph{NeurIPS}}. \bibinfo{pages}{6338--6347}.
\newblock


\bibitem[Nickel and Kiela(2018)]%
        {nickel2018learning}
\bibfield{author}{\bibinfo{person}{Maximillian Nickel} {and} \bibinfo{person}{Douwe Kiela}.} \bibinfo{year}{2018}\natexlab{}.
\newblock \showarticletitle{Learning Continuous Hierarchies in the Lorentz Model of Hyperbolic Geometry}. In \bibinfo{booktitle}{\emph{ICML}}. \bibinfo{pages}{3779--3788}.
\newblock


\bibitem[O'Shea and Nash(2015)]%
        {oshea2015introduction}
\bibfield{author}{\bibinfo{person}{Keiron O'Shea} {and} \bibinfo{person}{Ryan Nash}.} \bibinfo{year}{2015}\natexlab{}.
\newblock \showarticletitle{An Introduction to Convolutional Neural Networks}.
\newblock \bibinfo{journal}{\emph{arXiv:1511.08458}} (\bibinfo{year}{2015}).
\newblock


\bibitem[Pal et~al\mbox{.}(2025)]%
        {pal2025compositional}
\bibfield{author}{\bibinfo{person}{Avik Pal}, \bibinfo{person}{Max van Spengler}, \bibinfo{person}{Guido~Maria D'Amely~di Melendugno}, \bibinfo{person}{Alessandro Flaborea}, \bibinfo{person}{Fabio Galasso}, {and} \bibinfo{person}{Pascal Mettes}.} \bibinfo{year}{2025}\natexlab{}.
\newblock \showarticletitle{Compositional Entailment Learning for Hyperbolic Vision-Language Models}.
\newblock \bibinfo{journal}{\emph{ICLR}} (\bibinfo{year}{2025}).
\newblock


\bibitem[Papadopoulos et~al\mbox{.}(2010)]%
        {papadopoulos2010greedy}
\bibfield{author}{\bibinfo{person}{Fragkiskos Papadopoulos}, \bibinfo{person}{Dmitri Krioukov}, \bibinfo{person}{Mari{\'a}n Bogun{\'a}}, {and} \bibinfo{person}{Amin Vahdat}.} \bibinfo{year}{2010}\natexlab{}.
\newblock \showarticletitle{Greedy forwarding in dynamic scale-free networks embedded in hyperbolic metric spaces}. In \bibinfo{booktitle}{\emph{2010 Proceedings IEEE Infocom}}. IEEE, \bibinfo{pages}{1--9}.
\newblock


\bibitem[Peng et~al\mbox{.}(2021)]%
        {peng2021hyperbolic}
\bibfield{author}{\bibinfo{person}{Wei Peng}, \bibinfo{person}{Tuomas Varanka}, \bibinfo{person}{Abdelrahman Mostafa}, \bibinfo{person}{Henglin Shi}, {and} \bibinfo{person}{Guoying Zhao}.} \bibinfo{year}{2021}\natexlab{}.
\newblock \showarticletitle{Hyperbolic deep neural networks: A survey}.
\newblock \bibinfo{journal}{\emph{TPAMI}} (\bibinfo{year}{2021}).
\newblock


\bibitem[Pennec(2006)]%
        {pennec2006intrinsic}
\bibfield{author}{\bibinfo{person}{Xavier Pennec}.} \bibinfo{year}{2006}\natexlab{}.
\newblock \showarticletitle{Intrinsic Statistics on Riemannian Manifolds: Basic Tools for Geometric Measurements}.
\newblock \bibinfo{journal}{\emph{JMIV}} \bibinfo{volume}{25}, \bibinfo{number}{1} (\bibinfo{year}{2006}), \bibinfo{pages}{127--154}.
\newblock


\bibitem[Poleksic(2023)]%
        {poleksic2023hyperbolic}
\bibfield{author}{\bibinfo{person}{Aleksandar Poleksic}.} \bibinfo{year}{2023}\natexlab{}.
\newblock \showarticletitle{Hyperbolic matrix factorization improves prediction of drug-target associations}.
\newblock \bibinfo{journal}{\emph{Scientific Reports}} \bibinfo{volume}{13}, \bibinfo{number}{1} (\bibinfo{year}{2023}), \bibinfo{pages}{959}.
\newblock


\bibitem[Qu and Zou(2022a)]%
        {qu2022hyperbolic}
\bibfield{author}{\bibinfo{person}{Eric Qu} {and} \bibinfo{person}{Dongmian Zou}.} \bibinfo{year}{2022}\natexlab{a}.
\newblock \showarticletitle{Hyperbolic Neural Networks for Molecular Generation}.
\newblock \bibinfo{journal}{\emph{arXiv:2201.12825}} (\bibinfo{year}{2022}).
\newblock


\bibitem[Qu and Zou(2022b)]%
        {qu2022lorentzian}
\bibfield{author}{\bibinfo{person}{Eric Qu} {and} \bibinfo{person}{Dongmian Zou}.} \bibinfo{year}{2022}\natexlab{b}.
\newblock \showarticletitle{Lorentzian fully hyperbolic generative adversarial network}.
\newblock \bibinfo{journal}{\emph{arXiv:2201.12825}} (\bibinfo{year}{2022}).
\newblock


\bibitem[Radford et~al\mbox{.}(2021)]%
        {radford2021CLIP}
\bibfield{author}{\bibinfo{person}{Radford}, \bibinfo{person}{Kim Jong~Wook Alec}, \bibinfo{person}{Chris Hallacy}, \bibinfo{person}{Aditya Ramesh}, \bibinfo{person}{Gabriel Goh}, \bibinfo{person}{Sandhini Agarwal}, \bibinfo{person}{Girish Sastry}, \bibinfo{person}{Amanda Askell}, \bibinfo{person}{Pamela Mishkin}, \bibinfo{person}{Jack Clark}, \bibinfo{person}{Gretchen Krueger}, {and} \bibinfo{person}{Ilya Sutskever}.} \bibinfo{year}{2021}\natexlab{}.
\newblock \showarticletitle{Learning Transferable Visual Models From Natural Language Supervision}.
\newblock \bibinfo{journal}{\emph{arXiv:2103.00020}} (\bibinfo{year}{2021}).
\newblock


\bibitem[Rasley et~al\mbox{.}(2020)]%
        {rasley2020deepspeed}
\bibfield{author}{\bibinfo{person}{Jeff Rasley}, \bibinfo{person}{Samyam Rajbhandari}, \bibinfo{person}{Olatunji Ruwase}, {and} \bibinfo{person}{Yuxiong He}.} \bibinfo{year}{2020}\natexlab{}.
\newblock \showarticletitle{DeepSpeed: System Optimizations Enable Training Deep Learning Models with Over 100 Billion Parameters}. In \bibinfo{booktitle}{\emph{KDD}}. \bibinfo{pages}{3505--3506}.
\newblock


\bibitem[Said et~al\mbox{.}(2014)]%
        {said2014new}
\bibfield{author}{\bibinfo{person}{Salem Said}, \bibinfo{person}{Lionel Bombrun}, {and} \bibinfo{person}{Yannick Berthoumieu}.} \bibinfo{year}{2014}\natexlab{}.
\newblock \showarticletitle{New Riemannian Priors on the Univariate Normal Model}.
\newblock \bibinfo{journal}{\emph{Entropy}} \bibinfo{volume}{16}, \bibinfo{number}{7} (\bibinfo{year}{2014}), \bibinfo{pages}{4015--4031}.
\newblock


\bibitem[Sala et~al\mbox{.}(2018)]%
        {sala2018representation}
\bibfield{author}{\bibinfo{person}{Frederic Sala}, \bibinfo{person}{Chris De~Sa}, \bibinfo{person}{Albert Gu}, {and} \bibinfo{person}{Christopher Re}.} \bibinfo{year}{2018}\natexlab{}.
\newblock \showarticletitle{Representation Tradeoffs for Hyperbolic Embeddings}. In \bibinfo{booktitle}{\emph{ICML}}. \bibinfo{pages}{4460--4469}.
\newblock


\bibitem[Sanborn et~al\mbox{.}(2024)]%
        {sanborn2024beyond}
\bibfield{author}{\bibinfo{person}{Sophia Sanborn}, \bibinfo{person}{Johan Mathe}, \bibinfo{person}{Mathilde Papillon}, \bibinfo{person}{Domas Buracas}, \bibinfo{person}{Hansen~J. Lillemark}, \bibinfo{person}{Christian Shewmake}, \bibinfo{person}{Abby Bertics}, \bibinfo{person}{Xavier Pennec}, {and} \bibinfo{person}{Nina Miolane}.} \bibinfo{year}{2024}\natexlab{}.
\newblock \showarticletitle{Beyond Euclid: An Illustrated Guide to Modern Machine Learning with Geometric, Topological, and Algebraic Structures}.
\newblock \bibinfo{journal}{\emph{arXiv:2407.09468}} (\bibinfo{year}{2024}).
\newblock


\bibitem[Sarkar(2011)]%
        {sarkar2011low}
\bibfield{author}{\bibinfo{person}{Rik Sarkar}.} \bibinfo{year}{2011}\natexlab{}.
\newblock \showarticletitle{Low distortion delaunay embedding of trees in hyperbolic plane}. In \bibinfo{booktitle}{\emph{International Symposium on Graph Drawing}}. Springer, \bibinfo{pages}{355--366}.
\newblock


\bibitem[Shimizu et~al\mbox{.}(2020)]%
        {HNN++}
\bibfield{author}{\bibinfo{person}{Ryohei Shimizu}, \bibinfo{person}{Yusuke Mukuta}, {and} \bibinfo{person}{Tatsuya Harada}.} \bibinfo{year}{2020}\natexlab{}.
\newblock \showarticletitle{Hyperbolic Neural Networks++}. In \bibinfo{booktitle}{\emph{ICLR}}.
\newblock


\bibitem[Song et~al\mbox{.}(2021)]%
        {song2021scorebased}
\bibfield{author}{\bibinfo{person}{Yang Song}, \bibinfo{person}{Jascha Sohl-Dickstein}, \bibinfo{person}{Diederik~P Kingma}, \bibinfo{person}{Abhishek Kumar}, \bibinfo{person}{Stefano Ermon}, {and} \bibinfo{person}{Ben Poole}.} \bibinfo{year}{2021}\natexlab{}.
\newblock \showarticletitle{Score-Based Generative Modeling through Stochastic Differential Equations}. In \bibinfo{booktitle}{\emph{ICLR}}.
\newblock


\bibitem[Su et~al\mbox{.}(2021)]%
        {su2021roformer}
\bibfield{author}{\bibinfo{person}{Jianlin Su}, \bibinfo{person}{Yu Lu}, \bibinfo{person}{Shengfeng Pan}, \bibinfo{person}{Ahmed Murtadha}, \bibinfo{person}{Bo Wen}, {and} \bibinfo{person}{Yunfeng Liu}.} \bibinfo{year}{2021}\natexlab{}.
\newblock \showarticletitle{RoFormer: Enhanced Transformer with Rotary Position Embedding}.
\newblock \bibinfo{journal}{\emph{arXiv:2104.09864}} (\bibinfo{year}{2021}).
\newblock


\bibitem[Tang et~al\mbox{.}(2023)]%
        {tang2023hyperbolic}
\bibfield{author}{\bibinfo{person}{Xunzhu Tang}, \bibinfo{person}{Saad Ezzini}, \bibinfo{person}{Haoye Tian}, \bibinfo{person}{Yewei Song}, \bibinfo{person}{Jacques Klein}, \bibinfo{person}{Tegawende~F Bissyande}, {et~al\mbox{.}}} \bibinfo{year}{2023}\natexlab{}.
\newblock \showarticletitle{Hyperbolic code retrieval: a novel approach for efficient code search using hyperbolic space embeddings}.
\newblock \bibinfo{journal}{\emph{arXiv:2308.15234}} (\bibinfo{year}{2023}).
\newblock


\bibitem[Tifrea et~al\mbox{.}(2018)]%
        {tifrea2018poincar}
\bibfield{author}{\bibinfo{person}{Alexandru Tifrea}, \bibinfo{person}{Gary B{\'e}cigneul}, {and} \bibinfo{person}{Octavian-Eugen Ganea}.} \bibinfo{year}{2018}\natexlab{}.
\newblock \showarticletitle{Poincar\'e glove: Hyperbolic word embeddings}.
\newblock \bibinfo{journal}{\emph{arXiv:1810.06546}} (\bibinfo{year}{2018}).
\newblock


\bibitem[Ungar(2008)]%
        {ungar2008gyrovector}
\bibfield{author}{\bibinfo{person}{Abraham~Albert Ungar}.} \bibinfo{year}{2008}\natexlab{}.
\newblock \showarticletitle{A gyrovector space approach to hyperbolic geometry}.
\newblock \bibinfo{journal}{\emph{Synthesis Lectures on Mathematics and Statistics}} \bibinfo{volume}{1}, \bibinfo{number}{1} (\bibinfo{year}{2008}), \bibinfo{pages}{1--194}.
\newblock


\bibitem[van Spengler et~al\mbox{.}(2023)]%
        {van2023poincar}
\bibfield{author}{\bibinfo{person}{Max van Spengler}, \bibinfo{person}{Erwin Berkhout}, {and} \bibinfo{person}{Pascal Mettes}.} \bibinfo{year}{2023}\natexlab{}.
\newblock \showarticletitle{Poincar{\'e} ResNet}.
\newblock \bibinfo{journal}{\emph{CVPR}} (\bibinfo{year}{2023}).
\newblock


\bibitem[Vaswani et~al\mbox{.}(2017)]%
        {vaswani2017attention}
\bibfield{author}{\bibinfo{person}{Ashish Vaswani}, \bibinfo{person}{Noam Shazeer}, \bibinfo{person}{Niki Parmar}, \bibinfo{person}{Jakob Uszkoreit}, \bibinfo{person}{Llion Jones}, \bibinfo{person}{Aidan~N Gomez}, \bibinfo{person}{{\L}ukasz Kaiser}, {and} \bibinfo{person}{Illia Polosukhin}.} \bibinfo{year}{2017}\natexlab{}.
\newblock \showarticletitle{Attention is all you need}. In \bibinfo{booktitle}{\emph{NeurIPS}}. \bibinfo{pages}{5998--6008}.
\newblock


\bibitem[Villegas-Morcillo et~al\mbox{.}(2020)]%
        {villegas-morcillo2021protein}
\bibfield{author}{\bibinfo{person}{Amelia Villegas-Morcillo}, \bibinfo{person}{Stavros Makrodimitris}, \bibinfo{person}{Roeland C H~J van Ham}, \bibinfo{person}{Angel~M Gomez}, \bibinfo{person}{Victoria Sanchez}, {and} \bibinfo{person}{Marcel J~T Reinders}.} \bibinfo{year}{2020}\natexlab{}.
\newblock \showarticletitle{Unsupervised protein embeddings outperform hand-crafted sequence and structure features at predicting molecular function}.
\newblock \bibinfo{journal}{\emph{Bioinformatics}} \bibinfo{volume}{37}, \bibinfo{number}{2} (\bibinfo{year}{2020}), \bibinfo{pages}{162--170}.
\newblock
\showISSN{1367-4803}


\bibitem[Wang et~al\mbox{.}(2021)]%
        {wang2021dense}
\bibfield{author}{\bibinfo{person}{Xinlong Wang}, \bibinfo{person}{Rufeng Zhang}, \bibinfo{person}{Chunhua Shen}, \bibinfo{person}{Tao Kong}, {and} \bibinfo{person}{Lei Li}.} \bibinfo{year}{2021}\natexlab{}.
\newblock \showarticletitle{Dense contrastive learning for self-supervised visual pre-training}. In \bibinfo{booktitle}{\emph{CVPR}}.
\newblock


\bibitem[Wen et~al\mbox{.}(2024)]%
        {wen2024hyperbolic}
\bibfield{author}{\bibinfo{person}{Lingfeng Wen}, \bibinfo{person}{Xuan Tang}, \bibinfo{person}{Mingjie Ouyang}, \bibinfo{person}{Xiangxiang Shen}, \bibinfo{person}{Jian Yang}, \bibinfo{person}{Daxin Zhu}, \bibinfo{person}{Mingsong Chen}, {and} \bibinfo{person}{Xian Wei}.} \bibinfo{year}{2024}\natexlab{}.
\newblock \showarticletitle{Hyperbolic Graph Diffusion Model}. In \bibinfo{booktitle}{\emph{AAAI}}.
\newblock


\bibitem[Xie et~al\mbox{.}(2021)]%
        {xie2021unsupervised}
\bibfield{author}{\bibinfo{person}{Jiahao Xie}, \bibinfo{person}{Xiaohang Zhan}, \bibinfo{person}{Ziwei Liu}, \bibinfo{person}{Yew~Soon Ong}, {and} \bibinfo{person}{Chen~Change Loy}.} \bibinfo{year}{2021}\natexlab{}.
\newblock \showarticletitle{Unsupervised object-level representation learning from scene images}. In \bibinfo{booktitle}{\emph{NeurIPS}}.
\newblock


\bibitem[Yang et~al\mbox{.}(2024a)]%
        {yang2024hyplora}
\bibfield{author}{\bibinfo{person}{Menglin Yang}, \bibinfo{person}{Aosong Feng}, \bibinfo{person}{Bo Xiong}, \bibinfo{person}{Jihong Liu}, \bibinfo{person}{Irwin King}, {and} \bibinfo{person}{Rex Ying}.} \bibinfo{year}{2024}\natexlab{a}.
\newblock \showarticletitle{Hyperbolic Fine-tuning for Large Language Models}.
\newblock \bibinfo{journal}{\emph{ICML LLM Cognition Workshop}} (\bibinfo{year}{2024}).
\newblock


\bibitem[Yang et~al\mbox{.}(2022a)]%
        {yang2022hicf}
\bibfield{author}{\bibinfo{person}{Menglin Yang}, \bibinfo{person}{Zhihao Li}, \bibinfo{person}{Min Zhou}, \bibinfo{person}{Jiahong Liu}, {and} \bibinfo{person}{Irwin King}.} \bibinfo{year}{2022}\natexlab{a}.
\newblock \showarticletitle{Hicf: Hyperbolic informative collaborative filtering}. In \bibinfo{booktitle}{\emph{KDD}}. \bibinfo{pages}{2212--2221}.
\newblock


\bibitem[Yang et~al\mbox{.}(2024b)]%
        {yang2024hypformer}
\bibfield{author}{\bibinfo{person}{Menglin Yang}, \bibinfo{person}{Harshit Verma}, \bibinfo{person}{Delvin~Ce Zhang}, \bibinfo{person}{Jiahong Liu}, \bibinfo{person}{Irwin King}, {and} \bibinfo{person}{Rex Ying}.} \bibinfo{year}{2024}\natexlab{b}.
\newblock \showarticletitle{Hypformer: Exploring efficient transformer fully in hyperbolic space}. In \bibinfo{booktitle}{\emph{KDD}}. \bibinfo{pages}{3770--3781}.
\newblock


\bibitem[Yang et~al\mbox{.}(2021)]%
        {yang2021discrete}
\bibfield{author}{\bibinfo{person}{Menglin Yang}, \bibinfo{person}{Min Zhou}, \bibinfo{person}{Marcus Kalander}, \bibinfo{person}{Zengfeng Huang}, {and} \bibinfo{person}{Irwin King}.} \bibinfo{year}{2021}\natexlab{}.
\newblock \showarticletitle{Discrete-time Temporal Network Embedding via Implicit Hierarchical Learning in Hyperbolic Space}. In \bibinfo{booktitle}{\emph{KDD}}. \bibinfo{pages}{1975--1985}.
\newblock


\bibitem[Yang et~al\mbox{.}(2022b)]%
        {yang2022hyperbolicsurvey}
\bibfield{author}{\bibinfo{person}{Menglin Yang}, \bibinfo{person}{Min Zhou}, \bibinfo{person}{Zhihao Li}, \bibinfo{person}{Jiahong Liu}, \bibinfo{person}{Lujia Pan}, \bibinfo{person}{Hui Xiong}, {and} \bibinfo{person}{Irwin King}.} \bibinfo{year}{2022}\natexlab{b}.
\newblock \showarticletitle{Hyperbolic graph neural networks: a review of methods and applications}.
\newblock \bibinfo{journal}{\emph{arXiv:2202.13852}} (\bibinfo{year}{2022}).
\newblock


\bibitem[Yang et~al\mbox{.}(2022c)]%
        {yang2022hrcf}
\bibfield{author}{\bibinfo{person}{Menglin Yang}, \bibinfo{person}{Min Zhou}, \bibinfo{person}{Jiahong Liu}, \bibinfo{person}{Defu Lian}, {and} \bibinfo{person}{Irwin King}.} \bibinfo{year}{2022}\natexlab{c}.
\newblock \showarticletitle{{HRCF}: Enhancing Collaborative Filtering via Hyperbolic Geometric Regularization}. In \bibinfo{booktitle}{\emph{WebConf}}.
\newblock


\bibitem[Yang et~al\mbox{.}(2022d)]%
        {yang2022htgn}
\bibfield{author}{\bibinfo{person}{Menglin Yang}, \bibinfo{person}{Min Zhou}, \bibinfo{person}{Hui Xiong}, {and} \bibinfo{person}{Irwin King}.} \bibinfo{year}{2022}\natexlab{d}.
\newblock \showarticletitle{Hyperbolic Temporal Network Embedding}.
\newblock \bibinfo{journal}{\emph{TKDE}} (\bibinfo{year}{2022}).
\newblock


\bibitem[Yue et~al\mbox{.}(2023)]%
        {yue2023hyperbolic}
\bibfield{author}{\bibinfo{person}{Yun Yue}, \bibinfo{person}{Fangzhou Lin}, \bibinfo{person}{Kazunori~D. Yamada}, {and} \bibinfo{person}{Ziming Zhang}.} \bibinfo{year}{2023}\natexlab{}.
\newblock \showarticletitle{Hyperbolic Contrastive Learning}.
\newblock \bibinfo{journal}{\emph{arXiv:2302.01409}} (\bibinfo{year}{2023}).
\newblock


\bibitem[Zhang et~al\mbox{.}(2024)]%
        {zhang2024vision}
\bibfield{author}{\bibinfo{person}{Jingyi Zhang}, \bibinfo{person}{Jiaxing Huang}, \bibinfo{person}{Sheng Jin}, {and} \bibinfo{person}{Shijian Lu}.} \bibinfo{year}{2024}\natexlab{}.
\newblock \showarticletitle{Vision-Language Models for Vision Tasks: A Survey}.
\newblock \bibinfo{journal}{\emph{IEEE TPAMI}} \bibinfo{volume}{46}, \bibinfo{number}{8} (\bibinfo{year}{2024}), \bibinfo{pages}{5625--5644}.
\newblock


\bibitem[Zhang et~al\mbox{.}(2021a)]%
        {zhang2021hyperbolic}
\bibfield{author}{\bibinfo{person}{Yiding Zhang}, \bibinfo{person}{Xiao Wang}, \bibinfo{person}{Chuan Shi}, \bibinfo{person}{Xunqiang Jiang}, {and} \bibinfo{person}{Yanfang~Fanny Ye}.} \bibinfo{year}{2021}\natexlab{a}.
\newblock \showarticletitle{Hyperbolic graph attention network}.
\newblock \bibinfo{journal}{\emph{TBD}} (\bibinfo{year}{2021}).
\newblock


\bibitem[Zhang et~al\mbox{.}(2021b)]%
        {lgcn}
\bibfield{author}{\bibinfo{person}{Yiding Zhang}, \bibinfo{person}{Xiao Wang}, \bibinfo{person}{Chuan Shi}, \bibinfo{person}{Nian Liu}, {and} \bibinfo{person}{Guojie Song}.} \bibinfo{year}{2021}\natexlab{b}.
\newblock \showarticletitle{Lorentzian Graph Convolutional Networks}. In \bibinfo{booktitle}{\emph{WebConf}}. \bibinfo{pages}{1249--1261}.
\newblock


\bibitem[Zhang et~al\mbox{.}(2025)]%
        {zhang2025understanding}
\bibfield{author}{\bibinfo{person}{Yifei Zhang}, \bibinfo{person}{Hao Zhu}, \bibinfo{person}{Menglin Yang}, \bibinfo{person}{Jiahong Liu}, \bibinfo{person}{Rex Ying}, \bibinfo{person}{Irwin King}, {and} \bibinfo{person}{Piotr Koniusz}.} \bibinfo{year}{2025}\natexlab{}.
\newblock \showarticletitle{Understanding and mitigating hyperbolic dimensional collapse in graph contrastive learning}. In \bibinfo{booktitle}{\emph{KDD}}. \bibinfo{pages}{1984--1995}.
\newblock


\bibitem[Zhao et~al\mbox{.}(2023)]%
        {zhao2023survey}
\bibfield{author}{\bibinfo{person}{Wayne~Xin Zhao}, \bibinfo{person}{Kun Zhou}, \bibinfo{person}{Junyi Li}, \bibinfo{person}{Tianyi Tang}, \bibinfo{person}{Xiaolei Wang}, \bibinfo{person}{Yupeng Hou}, \bibinfo{person}{Yingqian Min}, \bibinfo{person}{Beichen Zhang}, \bibinfo{person}{Junjie Zhang}, \bibinfo{person}{Zican Dong}, {et~al\mbox{.}}} \bibinfo{year}{2023}\natexlab{}.
\newblock \showarticletitle{A survey of large language models}.
\newblock \bibinfo{journal}{\emph{arXiv:2303.18223}} \bibinfo{volume}{1}, \bibinfo{number}{2} (\bibinfo{year}{2023}).
\newblock


\bibitem[Zhu et~al\mbox{.}(2020)]%
        {zhu2020gil}
\bibfield{author}{\bibinfo{person}{Shichao Zhu}, \bibinfo{person}{Shirui Pan}, \bibinfo{person}{Chuan Zhou}, \bibinfo{person}{Jia Wu}, \bibinfo{person}{Yanan Cao}, {and} \bibinfo{person}{Bin Wang}.} \bibinfo{year}{2020}\natexlab{}.
\newblock \showarticletitle{Graph Geometry Interaction Learning}. In \bibinfo{booktitle}{\emph{NeurIPS}}, Vol.~\bibinfo{volume}{33}. \bibinfo{pages}{7548--7558}.
\newblock


\end{thebibliography}
\appendix

\section{Hyperbolic Geometry and Additional Works}\label{appendix:background}
\subsection{Hyperbolic Geometry} 
\textbf{Lorentz model.} An $n$-dimensional Lorentz model is
a Riemannian manifold $(\mathcal{L}^n, \mathfrak{g}_n^K)$ equipped with the Riemannian metric tensor $\mathfrak{g}_n^K = \mathrm{diag}(-1, 1, \ldots, 1)$ and defined by a constant negative curvature $K<0$, denoted as $\mathbb{L}^{K,n}$. Each point $\x\in\mathbb{L}^{K,n}$ has the parametrized form $[x_t, \x_s]^T$ where $x_t\in\R$ is called the time-like component and $\x_s\in\R^{n}$ is called the space-like component. $\mathbb{L}^{K,n}$ is equipped with the \textit{Lorentzian inner product}. For points $\x,\y\in\mathbb{L}^{K,n}$, their inner product $\langle\x,\y\rangle_\mathcal{L}$ is given by $\langle\x,\y\rangle_{\mathcal{L}} = -x_ty_t + \x_s^T\y_s = \x^T\mathfrak{g}_n^K\y,$
with $|\|\x\||_\mathcal{L}\coloneq\sqrt{|\langle \x, \x\rangle_\mathcal{L}|}$ being the Lorentzian norm. Formally, $\mathcal{L}^n$ is the point set $\mathcal{L}^n = \{\x\in\R^{n+1}: \langle\x,\x\rangle_\mathcal{L} = 1/K, x_t>0\}.$
The origin $\mathbf{o}\in\mathbb{L}^{K,n}$ is the point $[\sqrt{-1/K}, 0,\ldots,0]^T$. The tangent space at a point $\x\in\mathbb{L}^{K,n}$ is set of points orthogonal to $\x$, defined as
    $\mathcal{T}_\mathbf{x}\mathbb{L}^{K,n} = \{\y\in\R^{n+1}: \langle\x,\y\rangle_{\mathcal{L}} =0 \}$. 

\textbf{}{Poincar{\'e} aall model.} The Poincar{\'e} ball model of hyperbolic space, $\mathbb{P}^{K,n}$ is the $n$-dimensional sphere $S^n$ with radius $1/\sqrt{K}$, with the Riemannian metric $g_x^{\mathbb{P}} = \lambda_x^2 g^E$, where $ \lambda_x := \frac{2}{1 + K\|x\|^2}$ and $\mathfrak{g}^E$ is the Euclidean metric. The induced distance is given by $d_c(x,y) = \left( \frac{2}{\sqrt{-K}} \right) \tanh^{-1} \left( \sqrt{-K} \| -x \oplus_c y \| \right).$
\subsection{Hyperbolic Probabilistic Generative Models}
Some works adopted generative models to hyperbolic space, such as generative-adversarial networks (GAN)~\cite{lazcano2021hyperbolic, qu2022hyperbolic, li2023euclidean}. 
Here we introduce on hyperbolic probabilistic models, whose existing Euclidean versions are employed widely in generative foundation models.

\textbf{Hyperbolic Probability Distribution.} Normal distribution are vital for many probabilistic models~\cite{song2021scorebased, ho2020denoising, kingma2019introduction}. Prior works have adapted this to hyperbolic space through formulating wrapped hyperbolic normal distribution $\mathcal{N^\mathcal{M}}(\mathbf{\mu}, \Sigma)$~\cite{nagano2019wrapped, mathieu2019continuous}, where $\mathcal{M}$ is one of $\LL^{K,n}, \PP^{K,n}\}$, where one samples $\exp_\mathbf{\mu}^K\circ PT_{\oo\to\mathbf{\mu}}^K(\mathbf{v})$ where $\mathbf{v}\sim\mathcal{N}(0,\Sigma)$ over $\mathcal{T}_\oo \mathcal{M}$. The log density of this distribution is also easy to compute given the closed form geodesics~\cite{nagano2019wrapped, mathieu2019continuous}. Both works also proposed hyperbolic VAE models. Given dataset $D = \{\x_i\}_{i=1}^N$, the model trains a decoder $p_\theta(\x|\mathbf{z})$ that can create from $p_\theta=\mathcal{N}^\mathcal{M}(\mathbf{\mu}_0,\Sigma)$ a dataset that resembles $D$. Trained together is an encoder $q_\phi(\z|\x)$ by maximum the evidence lower bound (ELBO)\begin{equation}
    \log p_\theta(\x_i)\geq \mathbb{E}_{q_\phi(\z|\x_i)}\left[\log p_\theta(\x_i|\z)\right] - D_{KL}(q_\phi(\z|\x_i)\|p_\theta(\z))
\end{equation}
where $q_\phi = \mathcal{N}^\mathcal{M}(\mathbf{\mu}, \Sigma)$.

~\citet{mathieu2019continuous} also proposed using the Riemannian normal distribution~\cite{said2014new, pennec2006intrinsic}, together with a decoder that acknowledges the hyperbolic latent space structure using Riemannian Jensen's inequality in the ELBO loss, for the hyperbolic VAE model. 

\textbf{Hyperbolic Normalizing Flow.} Furthermore, HNF~\cite{bose2020latent} extended Euclidean normalizing flows to hyperbolic spaces, which consists of a series of functions $\varphi_i$ that transform the target distribution to a simple, known hyperbolic probability distribution. HNF gives two formulations of these functions. The first one implements $\varphi_i$ through $f^{\mathcal{T}, K}$ where $f$ is the Euclidean normalizing flow. The second defines the operation through an entirely hyperbolic function, by applying the logarithmic map and then parallel transporting between tangent spaces before lifting back to the manifold. 

RFM~\cite{chen2024flow} adapts the popular Euclidean flow matching~\cite{lipman2022flow} to general manifolds, which include hyperbolic space. Additionally, since hyperbolic geodesics have closed-form expressions, the algorithm is simulation-free and has a tractable form when applied to hyperbolic space using the maps $\exp_\x^K$ and $\log_\x^K$.  

\textbf{Hyperbolic Diffusion Models.} Diffusion models have been the state of the art for many generative tasks~\cite {croitoru2023diffusion}. RSGM~\cite{debortoli2022riemannian} and RDM~\cite{huang2022riemannian} adapted diffusion models~\cite{song2021scorebased, ho2020denoising} to Riemannian manifolds, which are applicable to hyperbolic space. For example, the score matching for $\HH^{K,n}$ from RSGM can be given as \begin{equation}
    L_{t}(\x_t, s_\theta) = \frac{1}{2}\mathbb{E}[\|s_\theta(\x_t)-\log_{\x_t}(\x_0)/t\|^2]
\end{equation} where $\x_t$ is the data at time $t$, obtained through simulation a manifold random walk. Specifically to hyperbolic spaces, a few studies have explored hyperbolic graph diffusion models through~\cite{wen2024hyperbolic, fu2024hyperbolic}. As a first attempt to define the hyperbolic graph diffusion model, HGDM~\cite{wen2024hyperbolic} defined the probability density at each time step $t$ similar to the Euclidean case~\cite{song2021scorebased, jo2022score}, replacing the Euclidean operations with its M{\"o}bius counterparts. 
For instance, the \textit{hyperbolic variance preserving SDE} has density \begin{equation}
    p(\x_t|\x_0) = \mathcal{N}^{\PP^{K,n}}\left(e^{-\frac{1}{2}\int_0^t\beta(s)ds}\otimes_K \x_0, I_n-I_ne^{-\int_0^t\beta(s)ds}\right)
\end{equation}
where $I_n$ is the identity and $\beta(s)$ is the same linear variance scheduler as~\cite{song2021scorebased}. 
HypDiff~\cite{fu2024hyperbolic} noticed that $\mathcal{N}^\mathbb{\mathcal{M}}$ noising process does not capture anisotropic properties in the latent space. They proposed a hyperbolic anisotropic diffusion framework that determines dominant diffusion directions by clustering nodes in hyperbolic latent space, considering both the radial and angular contributions.

\end{document}